%% file: main.tex
\documentclass[10pt,twocolumn,letterpaper]{article}

\usepackage[pagenumbers]{iccv_author_kit/iccv}      % To produce the REVIEW version

\input{math_commands}

\usepackage[utf8]{inputenc} % allow utf-8 input
\usepackage[T1]{fontenc}    % use 8-bit T1 fonts
\usepackage{url}            % simple URL typesetting
\usepackage{booktabs}       % professional-quality tables
\usepackage{amsfonts}       % blackboard math symbols
\usepackage{nicefrac}       % compact symbols for 1/2, etc.
\usepackage{microtype}      % microtypography
\usepackage{xcolor}         % colors
\usepackage{graphicx}
\usepackage{color, colortbl}
\usepackage{multirow}
\usepackage{pifont}% http://ctan.org/pkg/pifont
\usepackage{csquotes}
\usepackage{xspace}
\usepackage{siunitx}
\usepackage{caption} 
\usepackage{subcaption}

\definecolor{cvprblue}{rgb}{0.21,0.49,0.74}
\usepackage[pagebackref,breaklinks,colorlinks,citecolor=cvprblue]{hyperref}
\usepackage[capitalize]{cleveref}

\definecolor{Gray}{gray}{0.9}
\definecolor{mediumgray}{gray}{0.6}
\definecolor{darkgreen}{RGB}{0,180,0}
\definecolor{darkred}{RGB}{180,0,0}
\def\paperID{13838} % *** Enter the Paper ID here
\def\confName{ICCV}
\def\confYear{2025}

\title{Scene-Aware Location Modeling for Data Augmentation \\ in Automotive Object Detection}

\author{Jens Petersen\thanks{Equal contribution}\quad\;Davide Abati$^*$\quad  Amirhossein Habibian\quad\;Auke Wiggers\\
Qualcomm AI Research\thanks{\mbox{Qualcomm AI Research is an initiative of Qualcomm Technologies, Inc.}}\\
{\tt\small \{jpeterse,dabati,ahabibia,auke\}@qti.qualcomm.com}
}

\begin{document}
\maketitle

\interfootnotelinepenalty=10000

\input{sections/0_abstract}
\input{sections/1_introduction}
\input{sections/2_relatedwork}
\input{sections/3_method}

\input{sections/4_experiments}

\input{sections/5_conclusion}

{
    \small
    \bibliographystyle{iccv_author_kit/ieeenat_fullname}
    \bibliography{main}
}

% COMMENT OUT FOR SUBMISSION!!!
\input{sections/X_supplement}

\end{document}

%% file: math_commands.tex
\usepackage{amsmath,amsfonts,bm}

\def\xi{\mathbf{x}_i}

\def\z{\mathbf{z}}

\def\eqref#1{equation~\ref{#1}}
\def\1{\bm{1}}

\DeclareMathAlphabet{\mathsfit}{\encodingdefault}{\sfdefault}{m}{sl}
\SetMathAlphabet{\mathsfit}{bold}{\encodingdefault}{\sfdefault}{bx}{n}

%% file: sections/0_abstract.tex
\begin{abstract}
Generative image models are increasingly being used for training data augmentation in vision tasks.
In the context of automotive object detection, methods usually focus on producing augmented frames that look as realistic as possible, for example by replacing real objects with generated ones.
Others try to maximize the diversity of augmented frames, for example by pasting lots of generated objects onto existing backgrounds.
Both perspectives pay little attention to the locations of objects in the scene.
Frame layouts are either reused with little or no modification, or they are random and disregard realism entirely.
In this work, we argue that optimal data augmentation should also include realistic augmentation of layouts.
We introduce a scene-aware probabilistic location model that predicts where new objects can realistically be placed in an existing scene.
By then inpainting objects in these locations with a generative model, we obtain
much stronger augmentation performance than existing approaches.
We set a new state of the art for generative data augmentation on two automotive object detection tasks, achieving up to $2.8\times$ higher gains than the best competing approach ($+1.4$ vs. $+0.5$ mAP boost).
We also demonstrate significant improvements for instance segmentation.
\end{abstract}

%% file: sections/1_introduction.tex
\input{figures/cover}

\section{Introduction}
\label{sec:introduction}

\emph{Generative Data Augmentation} describes the use of generative models to create synthetic data that extends the training corpus of a learning model.
The appeal of \enquote{free} training data has long motivated related work \citep{chawla2002smote, he2008adasyn}, but with the recent progress in large generative image models \citep{ho2020denoising, rombach2022stablediffusion,imagen} the interest in this field has increased drastically, with promising successes in image classification~\cite{he2023synthetic,zhou_synthetic_2023} and object detection~\cite{xpaste,gen2det, dallefordet}.
This includes automotive scenes~\cite{geodiffusion, magicdrive}, the focus of this work, where the benefit of generative data augmentation is especially large, as edge case scenarios are often safety-critical and costly to acquire.
Existing methods for training data augmentation usually concern themselves with
improving the quality of generated objects, but they often neglect reasoning about their locations.
Some approaches reuse \emph{original} object locations from real frames~\cite{gen2det, magicdrive,kupyn2024dataset}, possibly with minor modifications \cite{geodiffusion,wang2024detdiffusion}, which results in augmented frames that are visual variations of the same scene. 
Alternatively, other methods add new objects in \emph{random} locations~\cite{xpaste}, completely ignoring the original scene composition, which results in unrealistic generations (\cref{fig:cover}).

In this work, we argue that object locations should also be considered a key component of data augmentation.
To demonstrate this, we propose a scene-aware probabilistic location model that, given an existing scene, predicts where a new object should be placed. 
Specifically, our model parses the scene to extract depth and drivable space, and it factorizes the joint probability of object categories, their locations, and their dimensions into a series of simpler conditional densities, which can be sampled from with ancestral sampling.
We then combine our location model with an inpainting diffusion model \cite{rombach2022stablediffusion} to render objects in the predicted locations, yielding augmented frames that are both realistic and different from existing scenes.
The result is a generative data augmentation technique that outperforms state-of-the-art approaches by a large margin, with a performance boost of up to $2.8\times$ w.r.t. the best competing approach ($+1.4$ vs. $+0.5$ mAP boost).
By modifying the inpainting model to produce both RGB and instance masks, we further demonstrate substantial gains in the instance segmentation setting.

\noindent In summary, our contributions are the following:
\begin{itemize}
    % \item We explore the effect of realistic location modeling in generative data augmentation, and find that it can have a large performance influence.
    \item We propose a scene-aware probabilistic location model that augments street scene layouts by placing new objects in realistic locations.
    % Placement is factorized into a series of simple likelihoods that are easy to approximate and sample from.
    \item We combine our location model with an inpainting diffusion model to produce augmented frames for object detector training, where our performance boost is up to $2.8\times$ higher compared to state-of-the-art approaches.
    \item By enabling the diffusion model to predict instance masks for generated objects, we further demonstrate substantial performance gains on instance segmentation.
\end{itemize}

%% file: figures/cover.tex
\begin{figure*}[t]
    \centering
    \includegraphics[width=0.93\linewidth]{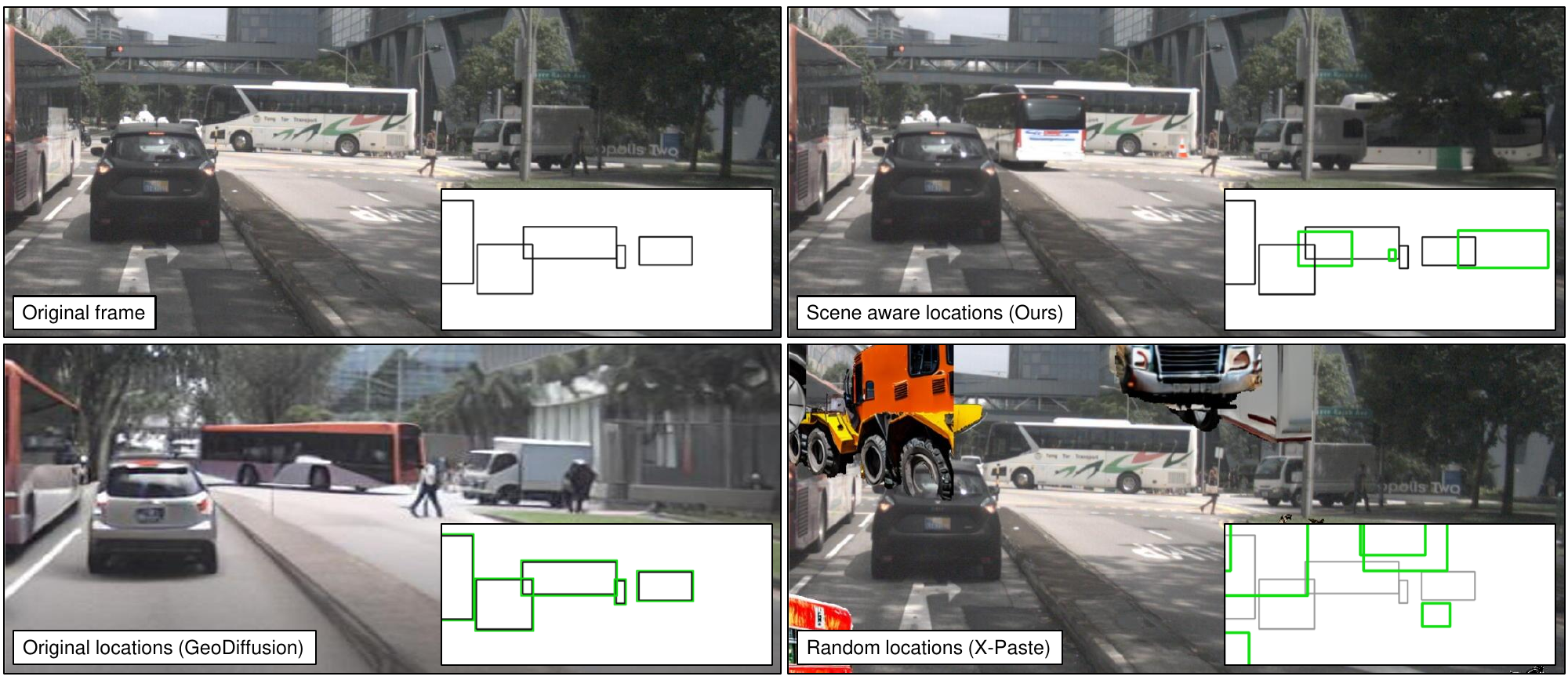}
    \caption{
An original scene (top left) and three augmented frames using different location modeling and augmentation strategies. 
Generated objects are indicated by green bounding boxes.
Our approach proposes locations that fit the original scene, resulting in novel compositions with high visual realism and challenging occlusion cases.
Approaches that reuse original locations, even with minor modifications such as in GeoDiffusion~\cite{geodiffusion}, generate frames with visual appearance diversity but limited location diversity.
Approaches that add objects in random locations such as X-Paste~\cite{xpaste} disregard the realism of the resulting layout and, in turn, of the generated frames.
}
\label{fig:cover}
\end{figure*}

%% file: sections/2_relatedwork.tex
\section{Related Work}
\label{sec:related}
%%%%%%%%%%%%%%%%%%%%%%%
\paragraph{Generative data augmentation.}
The use of synthetic instances to augment training data for vision tasks has recently become a common strategy~\cite{azizi2023synthetic,dallefordet,datasetdm,diffusionengine}.
% Creating realistic synthetic data typically required highly customized generative image models \cite{lee_contextaware_2018}.
% Indeed, the flexibility and adaptability of recent image generation models has led to renewed interest in generative data augmentation, achieving improvements on even long-standing benchmarks such as ImageNet~\citep{azizi2023synthetic}.
For object detector augmentation specifically, synthetic objects need to be rendered precisely in specified locations in a frame. 
Most works reuse object locations from real data~\citep{gen2det, kupyn2024dataset}, change only the background while leaving objects intact \cite{li2024simple},
or perform minor modifications such as translation or removal of bounding boxes~\citep{geodiffusion, magicdrive, magicdrive3d}.
The focus of these works is therefore mostly on improving the realism of generated data.
% Some follow-up work even examines whether the generated samples can be made more challenging~\citep{wang2024detdiffusion} for the downstream model.

Another popular augmentation strategy is ``cut-and-paste''~\cite{dwibedi2017cutpastelearn}: placing segmented or generated objects in real backgrounds in random locations.
Cut-and-paste approaches have shown to be very effective in both object detection and instance segmentation~\citep{dallefordet,xpaste,fan2024divergen,xie2024mosaicfusion}, despite the low realism of the resulting image.
% Recent approaches that place synthesized objects in random locations have also shown success in both object detection \citep{xpaste} and instance segmentation \citep{fan2024divergen, xie2024mosaicfusion} settings.
%despite the lack of realism of the augmented frames.
These results indicate that adding new objects in new locations may be just as important as creating realistic images, echoing earlier findings~\cite{dvornik2019importancevisualcontext}.
Our work achieves both by adding \emph{new} objects in \emph{realistic} locations to augment frames.
% This reaffirms findings by the seminal work of \citet{dvornik2019importancevisualcontext} that shows the importance of realistic locations for cut-and-paste augmentation approaches.
% We directly inpaint with a powerful generative model, leading to higher realism of augmented frames.
% However, whereas they paste segmented objects on existing backgrounds, we directly render synthetic targets in context by using powerful inpainting models, further improving the realism of augmented frames. 

%
%%%%%%%%%%%%%%%%%%%%%%%
\paragraph{Layout generation and location modeling.}
Predicting object locations is related to layout completion and generation.
Dedicated methods typically solve these tasks by modeling interactions at the bounding box level without additional context \citep{jyothi2019layoutvae, yang2021layouttransformer, gupta2021layouttransformer, kong2022blt, inoue2023layoutdm}, and are usually applied to design documents or other highly structured data.
% However, these methods only reason about relative positions of bounding boxes.
%Differently, 
% When adding synthetic objects to an existing background, visual cues from the scene should also be taken into account to determine realistic object scale and location.
Approaches that take scene information into account to determine object locations do exist~\cite{yun2024generative, topnet}, but often require paired training datasets of (empty) images and feasible object placements, which are not readily available for automotive scenes.
They may also require an image of the segmented object to place~\cite{topnet, placenet}, or have only been shown to work for specific object categories such as cars and pedestrians \citep{lee_contextaware_2018}.
In this work, we instead want to determine the location before such an object is available.
%To avoid the need for paired data for supervision, we factorize placement into a series of simple, easy to model likelihoods.
% We therefore opt for a factorized probabilistic model that takes scene information into account at a high level through depth maps and semantic information, thus avoiding the need to create a paired dataset.
%Some works focus on modeling plausible actor placement in street scenes.
%Similar to the aforementioned methods, they either require segmented objects to be placed~\cite{placenet}, or have only been demonstrated for specific object categories, such as cars and pedestrians~\cite{lee_contextaware_2018}. 
Finally, some approaches reason about object locations in 3D space~\cite{metasim,metasim2,scenegen}, but this requires detailed 3D annotations, which are usually much harder to obtain than 2D annotations.
% Recent work also uses 3D annotations for generative data augmentation of 3D multi-view detectors~\cite{magicdrive,magicdrive3d,magicdrivedit}, but we restrict ourselves to 2D detector augmentation here, as that is still the dominant approach in literature.

% \jp{Move this somewhere else}

%% file: sections/3_method.tex
\input{figures/model}

\section{Method}
\label{sec:method}

Our goal is to augment street scenes by determining where new objects can be placed.
Using these locations, we augment frames for detector training, as illustrated in \cref{fig:pipeline}.
We first describe our proposed scene-aware location model, a probabilistic approach that factorizes the likelihood of new object locations into a sequence of simple conditional likelihoods.
We then describe our strategy to render these objects into the existing scene to obtain augmented frames.

\subsection{A factorized scene-aware location model}
\label{sub:method:locationmodel}

Each generated object is described by a class label $c$ and a 2D bounding box $b$ specifying its location and dimensions in the given scene.
The procedure for placing a new object into an existing scene can be thought of as a two-step process: 
1) decide \emph{what} object to place and 
2) determine \emph{where} to place it (and what size it should have).

\input{figures/box_location_model}
\input{figures/box_location_examples}

We make these choices explicit in a likelihood model $\hat{p}$ that approximates the true probability density $p$ of object categories, locations and scales.
Since the distribution of plausible object placements is highly dependent on the constraints imposed by the given scene, we condition our model on high level scene descriptions.
Specifically, we assume to have scene representations in the form of a depth map $\mathbf{D}$ and a \emph{drivable space} semantic map\footnote{We define semantic categories \enquote{road}, \enquote{terrain}, \enquote{sidewalk} to be drivable space, as this is where objects of interest typically appear.} $\mathbf{S}$.
The depth map tells us something about the structure of the 3D scene and distance to existing objects, while the semantic map tells us exactly where the ground plane is.

We are now interested in predicting plausible objects and their locations.
We approximate the likelihood of class $c$, the distance to the object $d$, and bounding box center, height and width $(b^x,b^y,b^w,b^h)$ by factorizing it as follows:
\begin{align}
\label{eq:location_model}
& p(c,b^x,b^y,b^w,b^h,d \, | \, \mathbf{D},\mathbf{S}) \approx \nonumber\\
&\,\,\, \hat{p}(b^w|b^h,c) \, \hat{p}(b^h|d,c) \, \hat{p}(b^x,b^y|d,\mathbf{D},\mathbf{S}) \, \hat{p}(d|c) \, \hat{p}(c),
\end{align}
where $d$ is a sampled depth value, used only as an intermediate variable.
The corresponding graphical model is visualized in \cref{fig:box_location_model}.
This factorization is chosen such that individual terms in \cref{eq:location_model} are easy to approximate with empirical or simple parametric distributions.
% We refer the reader to the supplementary material for a more detailed analysis of these conditionals.
Using these approximations, we can use ancestral sampling to generate realistic object location proposals for the scene:
\begin{enumerate}
\item \textbf{Sample a class.}
We first sample a class from the multinomial
$\hat{p}(c)$, which we choose to have uniform probabilities to oversample rare classes.
%relative to the true class distribution.
\item \textbf{Sample a depth.} 
To sample objects at realistic distances, we collect observed object depths per class from training data, and approximate $p(d|c)$ with a log-normal distribution. 
%This ensures the object location is at a reasonable camera distance.
In comparison, we observed that directly choosing a random location in the drivable space would result in oversampling of objects at short distances.
\item \textbf{Sample a location.} Using semantic map $\mathbf{S}$ and depth map $\mathbf{D}$ we select $b^x,b^y$ uniformly at random from the scene's drivable space, limited to locations with depths that are within a threshold $\tau_d$ to the sampled distance $d$. \cref{fig:box_location_model} shows examples of such ``placement bands''.
\item \textbf{Sample a height.} 
We collect statistics of object heights at different depth intervals, and approximate them with log-normal distributions $\hat{p}(b^h|d,c)$.
\item \textbf{Sample a width.} We collect aspect ratios of objects, independent of depth, and use the resulting empirical distributions (\ie the histograms) for sampling object widths $b^w \sim \hat{p}(b^w|b^h,c)$. 
\end{enumerate}
Examples of boxes generated by our model are shown in \cref{fig:box_location_examples}.
We provide more details on the steps and evaluate the quality of the approximations in the supplementary material. 
\subsection{Generative augmentation}
\label{sub:method:gendataaug}
In the previous section we introduced a 
%training-free 
probabilistic location model that places new objects, parametrized by class and bounding box, into an existing scene (\cref{fig:pipeline}~(A)).
In order to render the desired objects, 
we use a diffusion model for inpainting, namely Stable Diffusion 2 (SD2)\footnote{\url{https://huggingface.co/stabilityai/stable-diffusion-2-inpainting}}~\cite{rombach2022stablediffusion}.
As operating at high resolution with diffusion models is not straightforward, we extract square patches centered in the proposed locations.
Every patch has a resolution of $m \times m$ pixels, where $m=2\times max(b^h,b^w)$, and is resized to a fixed resolution of $512\times512$ for inpainting.
Examples of such patches are shown in \cref{fig:pipeline}~(B).
Prior work utilizes large language models to craft complex textual descriptions~\citep{datasetdm}, but we found that simple text prompts in the format\,\,``\texttt{image of a <class name>}'' are sufficient for realistic generations.
We finetune the diffusion model on the domain of interest, for which we tried both direct finetuning and ControlNet~\cite{controlnet}. 
We report scores with ControlNet, but the two options perform on par (see Appendix).
Finetuning benefits the inpainting model in three ways.
First, it allows it to adapt to the pixel-level statistics of the target dataset, \ie to generate objects that look natural in terms of saturation and contrast.
Second, it resolves ambiguities in textual category labels: for example, the class \enquote{rider} can be interpreted by SD2 as \enquote{horse rider}, whereas in the BDD100K dataset it represents only \enquote{motorcycle riders}. 
Third, finetuning forces objects to fit more tightly in the provided bounding box.
We show examples of this in the Appendix.
\input{figures/examples}
\paragraph{Obtaining object masks.}
To augment data for instance segmentation tasks, we need instance masks for every synthetic object.
To this end, we equip the inpainting model with a simple mask decoding module $\mathcal{M}$, responsible for providing a segmentation mask for the objects it generates.

The mask decoder $\mathcal{M}$ is created as a lightweight copy of the SD2 UNet-decoder, with 4x fewer channels per layer.
It receives multi-scale features from the SD2 UNet-encoder as input, as its representations are rich in semantic information about objects being generated~\citep{datasetdm,diffusionengine}.
Specifically, whenever generating an object, we pass the \enquote{denoised} latent variable $\z_0$ to the UNet and extract representations $\{ \mathbf{r}_1, \mathbf{r}_2, \dots \mathbf{r}_d \}$ at multiple resolutions, before each downsampling layer in the architecture.
These features then undergo a simple multi-scale aggregation phase, before being fed to $\mathcal{M}$ to decode an alpha mask $\hat{s}$.
To train $\mathcal{M}$, we assume access to crops with available groundtruth instance masks $s$ and optimize a simple binary cross-entropy loss.
More details are given in the Appendix.
%
% \begin{equation}
% \label{eq:mask}
% \hat{s} = \mathcal{M}(\mathbf{r}_1, \mathbf{r}_2, \dots \mathbf{r}_d), \quad
% \mathcal{L}_{mask} =\mathbb{E} \|s - \hat{s} \|_1.
% \end{equation}
%
% We find that using a regression loss instead of a segmentation loss allows the masks to have softer edges, which leads to better forground-background blending.

Besides enabling instance segmentation augmentation, we found that these masks 
allow more realistic handling of occlusions between generated objects, without artifacts (see~\cref{fig:pipeline} (C), occlusion between bus and barrier).
Moreover, it allows us to refine the bounding box size in the case the inpainting model generates an object that is smaller than the input bounding box.
We provide more details on mask decoding in the supplementary material.

%% file: figures/model.tex
\begin{figure*}[t]
\centering

\includegraphics[width=0.95\textwidth]{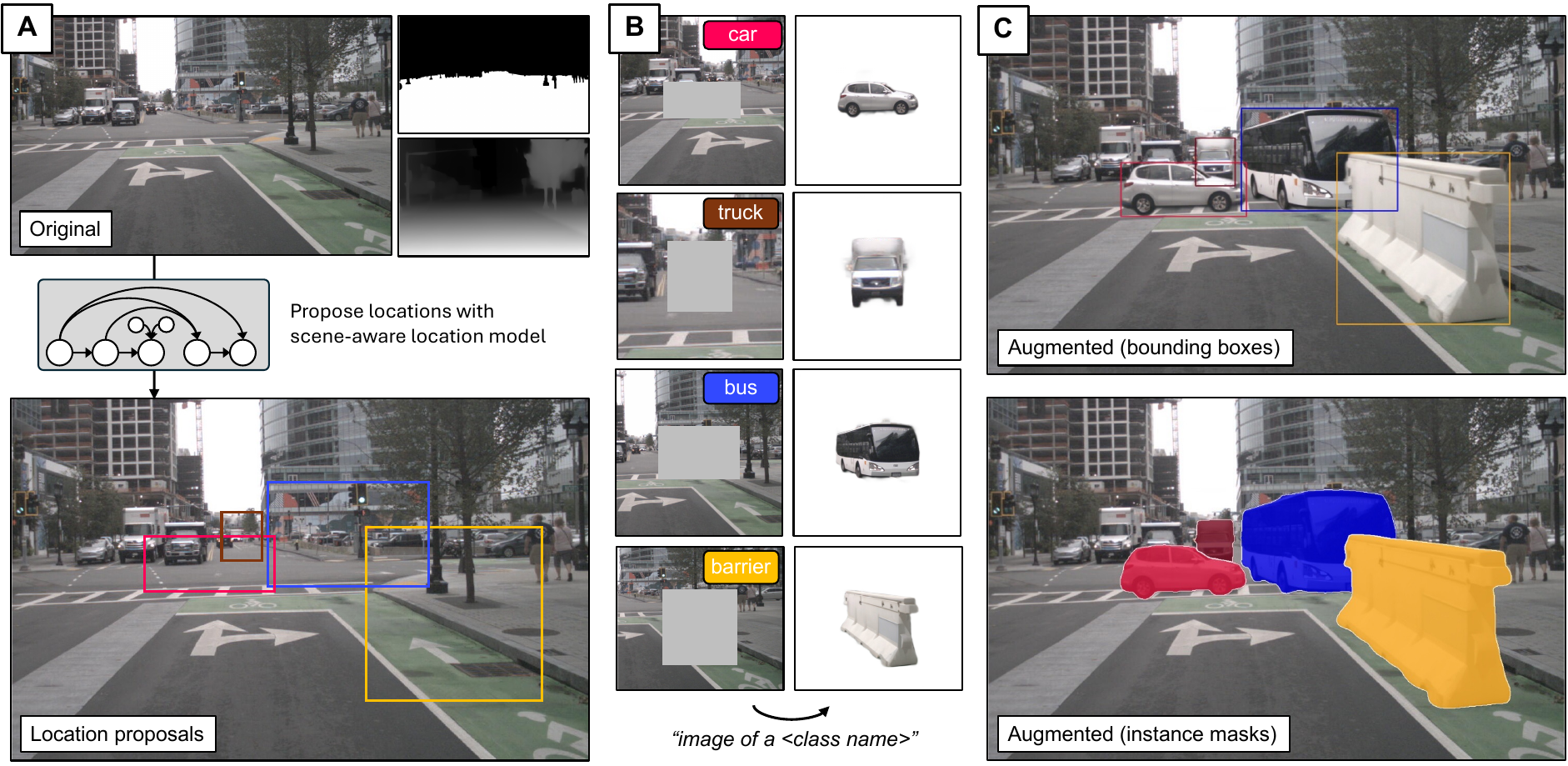}
\caption{
Overview of our augmentation pipeline. 
(A) We first use the location model to predict realistic bounding box locations for new objects, using depth and drivable space segmentation. 
%for realistic locations and scales. 
(B) We then generate an object and corresponding instance mask using an inpainting model.  
(C) This allows us to create pseudo-labels for object detection and instance segmentation. 
Our approach scales to high resolution images, and creates realistic and challenging occlusion cases.
%(only annotations for generated objects are shown). 
%By treating objects separately we can include or exclude each to create diverse augmented realizations of the same frame, as seen in FIGURE}
}
\label{fig:pipeline}
\end{figure*}

%% file: figures/box_location_model.tex
\begin{figure}
    \centering
    \includegraphics[width=0.95\linewidth]{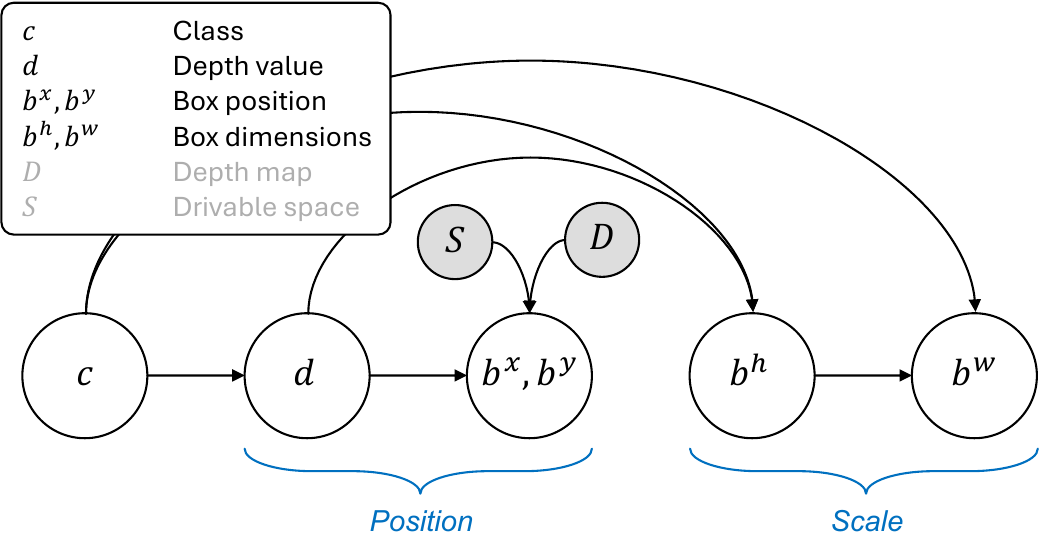}
    \includegraphics[width=\linewidth]{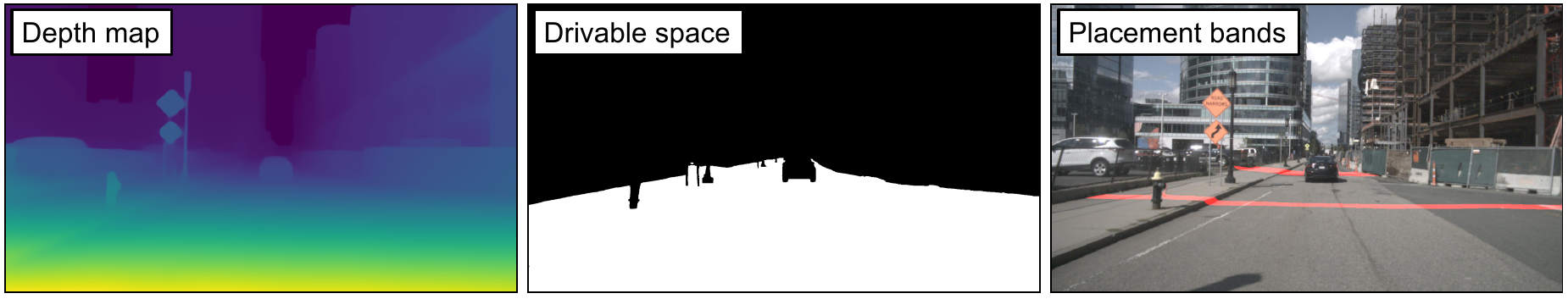}
    \caption{(Top) Our location model factorizes object placement into a series of conditional likelihoods, each of which is easy to approximate or parametrize. (Bottom) We sample a desired distance to the object, $d$, and determine admissible locations for this depth (red lines are two separate examples of such \emph{placement bands}).}
    \label{fig:box_location_model}
\end{figure}

%% file: figures/box_location_examples.tex
%%  PORTRAIT - ONE COLUMN IMAGE
% \begin{figure}[t]
%     \centering
%     % \includegraphics[width=\linewidth]{img/locations/location_portrait_1.pdf}
%     % \includegraphics[width=\linewidth]{img/locations/location_portrait_2.pdf}
%     % \includegraphics[width=\linewidth]{img/locations/location_portrait_3.pdf}
%     \caption{Examples of boxes generated by our model (top) and by random placement following XPaste \cite{xpaste} (bottom)}
%     \label{fig:box_location_examples}
% \end{figure}
%
%%  LANDSCAPE - ONE COLUMN IMAGE
\begin{figure*}[t]
\centering
\includegraphics[width=0.9\textwidth]{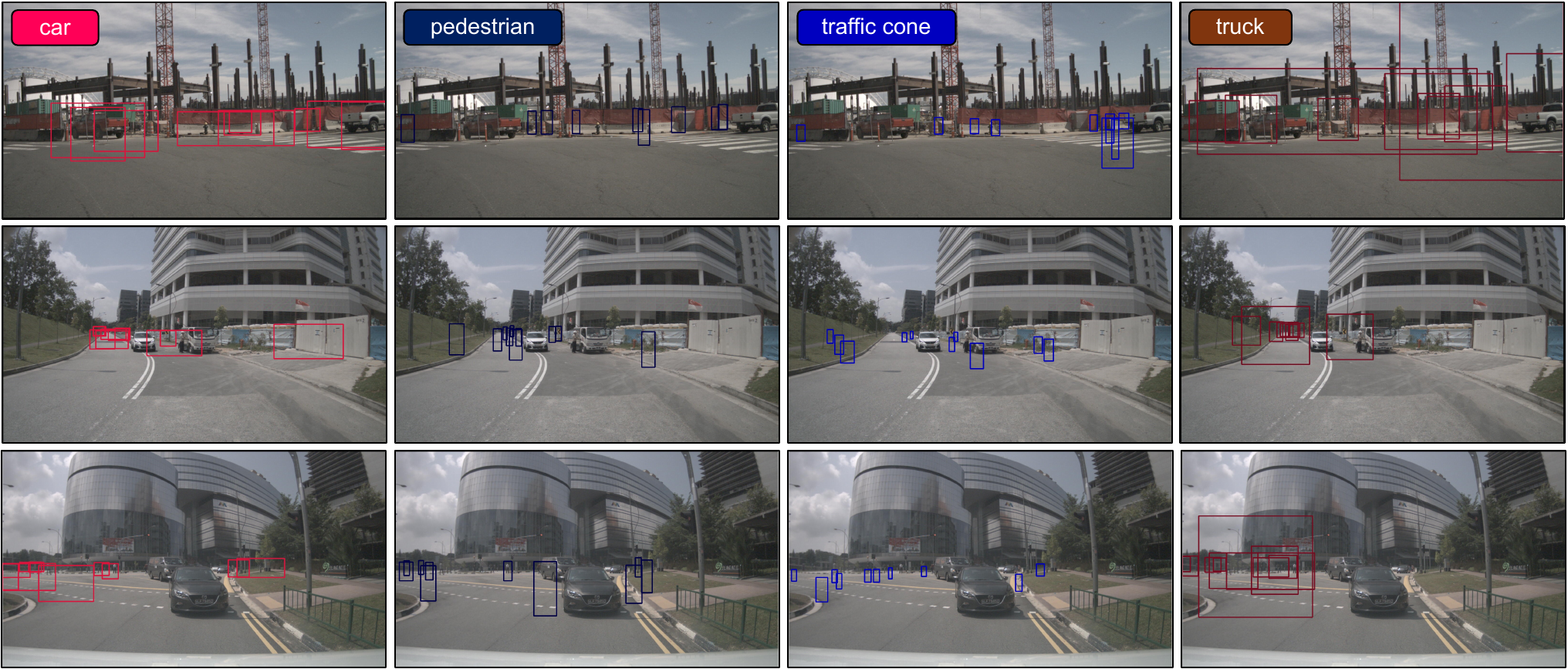}
\caption{Example bounding box proposals from our location model, separated by class.}
\label{fig:box_location_examples}
\end{figure*}

%% file: figures/examples.tex
\begin{figure*}[tbh]
    \centering
    \includegraphics[width=\linewidth]{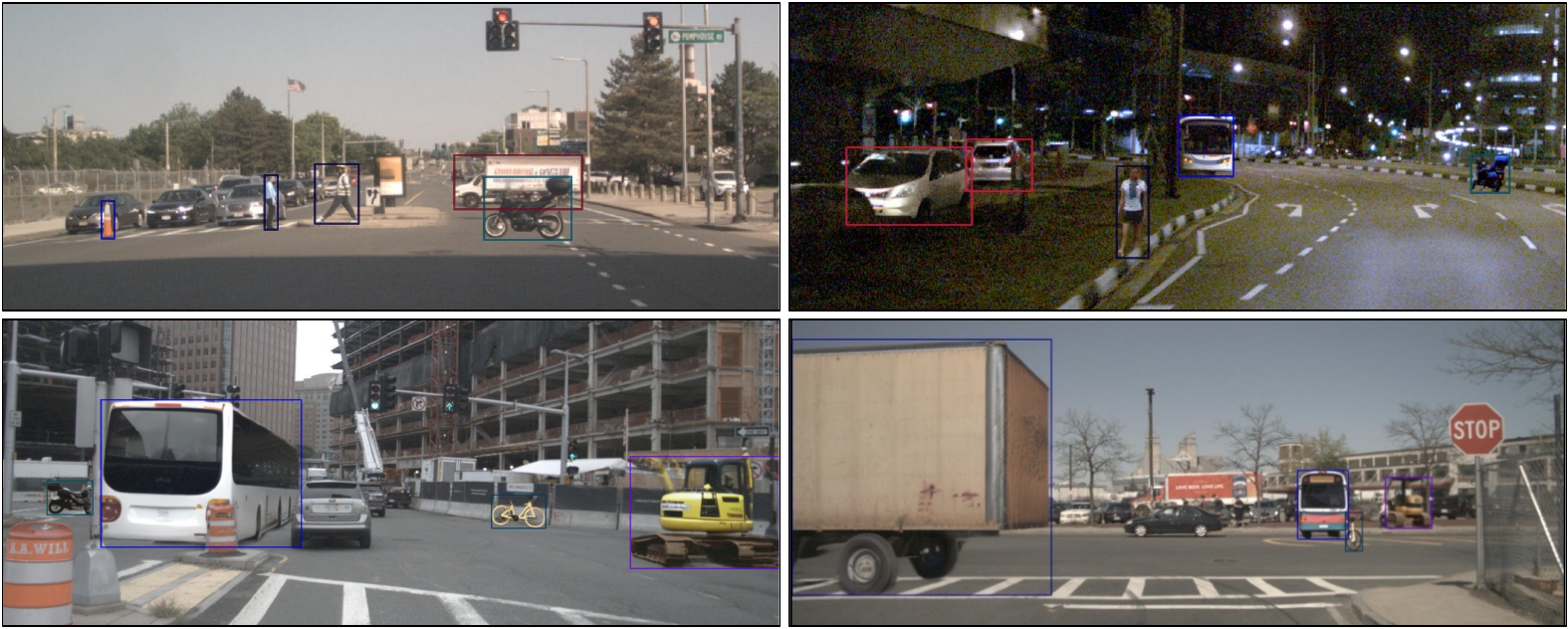}
    \caption{
    Example of nuImages frames augmented with our approach. 
    We show the bounding boxes for all added objects.
    In diverse scenarios, the location and scale of added objects are realistic and thus result in realistic augmented images.
    }
    \label{fig:examples}
\end{figure*}

%% file: sections/4_experiments.tex
\section{Experiments}
\label{sec:experiments}
\input{tables/sota_nuimages}
In this section we present experimental results for generative data augmentation.
We show main experiments in \cref{sub:train_data_aug}, where we augment data for both object detection and instance segmentation.
We then investigate the influence of different design choices in more detail in \cref{sub:ablations}.
% In \cref{sub:realism_diversity} we analyze the realism and diversity of generated data, which we believe are key aspects for successful training data augmentation. 
% Finally, we ablate some design choices in \cref{sub:ablations}.
%
%%%%%%%%%%%%%%%
\paragraph{Datasets and evaluation.} 
We conduct experiments on two public automotive object detection benchmarks: nuImages~\citep{caesar2020nuscenes} and BDD100K~\citep{yu2020BDD100K}.
The nuImages dataset contains \num{67,279} training images and \num{16,445} validation images, at resolution $1600 \times 900$.
It is published by Motional AD Inc. under a CC BY-NC-SA 4.0 license.
The BDD100K dataset contains \num{70,000} training and \num{10,000} validation images, at resolution $1280 \times 720$.
To compare data augmentation strategies, we always use all available real training images and equally many augmented frames during each training epoch.
Following standard practice, we evaluate object detectors through the mean Average Precision (mAP) on real validation images.
For object detection experiments we use a Faster R-CNN~\cite{fasterrcnn} with a ResNet-50 backbone (pretrained on ImageNet~\cite{imagenet}).
For instance segmentation experiments, we use a Mask R-CNN~\cite{maskrcnn} with the same backbone.
%
%%%%%%%%%%%%%%%
\paragraph{Baselines.}
To measure the effectiveness of our location model, we compare against two baselines that use the same generator (as described in \cref{sub:method:gendataaug}), but use a different placement strategy.
The first baseline re-uses the original object locations, \ie it generates instances in existing locations, effectively replacing original objects with synthetic ones.
This approach is similar to prior work on object detector augmentation~\cite{gen2det,kupyn2024dataset}.
The main differences are that Gen2Det~\cite{gen2det} uses an unspecified closed-source diffusion model, whereas \citet{kupyn2024dataset} use additional depth and edge conditioning for generation. 
We refer to this baseline as \enquote{Replacement} in the following.
The second baseline adds new objects in random locations, independently of the scene, which we call \enquote{Random Loc.}.
For the latter, bounding box \emph{sizes} follow the training distribution, while the \emph{locations} are uniformly sampled in the frame.

Additionally, we compare to two state-of-the-art augmentation methods.
X-Paste~\citep{xpaste}---a reference work in cut-and-paste augmentation---generates synthetic objects with Stable Diffusion~\cite{rombach2022stablediffusion}, segments them, and then pastes them on an existing frame with random location and scale.
We use the publicly released code\footnote{\href{https://github.com/yoctta/XPaste}{\url{https://github.com/yoctta/XPaste}}} to generate \num{100,000} objects, and we paste up to \num{10} per frame, selected at random.
We further compare to results reported by GeoDiffusion~\cite{geodiffusion}, a layout-to-image method that renders synthetic frames from slightly perturbed object locations.
For this model, we report metrics from the original publication, as the authors only released inference code and lower-resolution models.
Finally, we also reimplemented background augmentation from \cite{li2024simple}, but only observed negative augmentation performance, so we provide those results in the supplementary material.

%
%%%%%%%%%%%%%%%%%%
\paragraph{Implementation details.}
We augment frames by generating 12 new objects per frame, and randomly showing each with 0.5 probability whenever the frame is chosen during training.
% Early ablations showed that generating more objects per frame did not result in higher performance.
For the inpainting model, we finetune for 300,000 iterations at a batchsize of 16, using ControlNet~\cite{controlnet} with masked crops as conditioning input.
All detector trainings are performed using the mmdetection library~\citep{mmdetection} and the default configurations, except for the number of training epochs that is set to 36 for all datasets and models.
To enable a fair comparison to GeoDiffusion on nuImages, we follow the protocol in the original paper and further train (and evaluate) the detector at a reduced resolution ($800\times456$ pixels);
for this experiment exclusively, we reduce the number of epochs to 12 to match their setup.
When training on BDD100K we do not augment the \emph{traffic sign} and \emph{traffic light} categories, as our location model is better suited for objects on the ground.
The mAP is however computed on all classes.
For our scene-aware location model, we extract scene representations from off-the-shelf models for depth estimation~\cite{depthanything} and semantic segmentation~\cite{yang2023dense}.
We train our mask decoder on nuImages, as the dataset provides precise instance masks.
%
%%%%%%%%%%%%%%%%%%%%%%%%%%%%%%%%%%%%%%%%%%%%%%
\subsection{Training data augmentation}
\label{sub:train_data_aug}
\input{tables/masks_nuimages}
\input{tables/sota_bdd}
\paragraph{Object detection on nuImages.}
We first analyze the performance of different augmentation strategies for object detection on nuImages, for which we report both mAP and class specific scores in \cref{tab:sota_nuimages}. 
Examples of augmented frames from our method are shown in \cref{fig:examples}.
Although all augmentation methods improve over the baseline (trained on real data only), we can make the following observations.

At low resolution ($800\times456$), methods using random locations improve mAP marginally (up to +0.4 points), whereas techniques leveraging original locations prove more successful (up to +0.7 points).
This is especially true for classes like \emph{pedestrian}, \emph{cone} and \emph{barrier}, which are generally smaller than other objects and therefore harder to detect: diversifying their appearance while keeping the location unchanged clearly helps detector training.
However, this observation is reversed at full resolution, where the replacement strategy underperforms with respect to random locations.
In both cases, our augmentation method proves to be the best approach, significantly increasing the mAP of the detector by 1.4 and 1.6 in low and high resolution, respectively.

Looking at per-category results, we can see how our approach improves the most on classes that are under-represented in the dataset, such as \emph{trailer} ($0.6\%$ of training instances), \emph{construction vehicle} ($0.9\%$), \emph{bus} ($1.2\%$), \emph{bicycle} ($2.5\%$) and \emph{motorcycle} ($2.5\%$).
In contrast, strategies relying on original object locations (\eg replacement) tend to work well on categories that are already well represented in the training data, such as \emph{car} and \emph{pedestrian}. 
This is likely due to the fact that they cannot easily oversample rare classes, unlike our method and approaches that use random locations.
This finding suggests that approaches that allow oversampling of rare classes, such as ours, are the superior choice for problems where categories follow a long-tailed distribution. 
Arguably, this applies to many real-world problems.
%
%%%%%%%%%%%%%%%%%%
\paragraph{Instance segmentation on nuImages.}
Next, we test our approach on instance segmentation, by using the strategy described in \cref{sub:method:gendataaug} to obtain pseudo-groundtruth masks for synthetic objects.
We compare our method to replacement and random locations by assessing mAP both on bounding boxes and on instance masks, and we report results in \cref{tab:masks_nuimages}.
Our approach outperforms both baselines in both metrics, highlighting the benefit of scene-aware location modeling. 
The baselines differ from our approach only in the chosen object locations, while the generation model is the same.

We also observe that the performance on small objects remains largely unchanged, regardless of the augmentation method.
Moreover, the improvement over the baseline offered by our method seems to increase with object size.
A potential reason for this behavior is that small objects are more common in the real data, leaving less room for improvement compared to the relatively rare large objects.
The bottom row of \cref{tab:sota_nuimages} shows the distribution of object counts in the data.
\paragraph{Object detection on BDD100K.}
We repeat the object detection augmentation experiment on the BDD100K dataset, and report results in \cref{tab:sota_bdd}.
Overall, the table shows lower scores than on nuImages, and we ascribe this behaviour to BDD100K being a more challenging dataset (\eg it includes data filmed through the windshield and thus it shows a lot of reflections and dirt).
Scene-aware location modeling again outperforms both replacement and random location strategies, improving the baseline detector by \num{1.3} points in mAP ($31.4\rightarrow 32.7$). 

X-Paste also performs reasonably well on this dataset, and it notably outperforms our method in the \emph{train} category.
We believe this result is due to the scarcity of data for this class, which features only 15 instances in the validation set, and for which the AP is extremely low for all methods.
We suspect that in such extreme long-tailed cases the high diversity of generated objects offered by X-Paste might prove more beneficial than realistic placement and generation.

\subsection{Ablations}
\label{sub:ablations}

We use this section to investigate the influence of individual model components and design choices. The results are compiled in \cref{tab:ablation}, we address them one-by-one.

\paragraph{Finetuning.}

First, we ablate the decision to finetune the inpainting model on the target dataset, for which we explored both direct finetuning and ControlNet~\cite{controlnet} (see Appendix for a comparison).
While the SD2 base model sometimes creates convincing objects, we find that on average, finetuning leads to a) better visual coherence with the surroundings and b) objects that better fill the provided bounding box.
We show visual examples in the Appendix.
% We show examples of this in the appendix, and mAP results for Faster R-CNN training on nuImages are shown in \cref{tab:ablation}.
Consequently, performance without finetuning is much lower, and hardly improves over the baseline model.
We leave an exploration of other finetuning techniques~\cite{lora,t2i} for future work.

\paragraph{Mask prediction, SAM masks.}

At inference time, we generate objects and masks jointly.
This is not strictly necessary for detector augmentation, and mainly serves to extend our approach to instance segmentation.
However, we can also use the generated masks to refine the bounding boxes and improve foreground-background blending.
The effect is strongest at high IoU thresholds (see Appendix), but it significantly influences mAP as a whole.
Interestingly, using SAM~\cite{sam} to extract segmentation masks for generated objects works only slightly less well at low resolution, but leads to a large performance drop at full resolution.
We suspect that even though SAM was trained for broad applicability, there is still a distribution mismatch, and our model benefits from nuImages training. 
While this could be remedied with finetuning SAM, our mask decoder is orders of magnitude smaller than it (we use the \emph{sam-vit-huge} checkpoint here).
This result highlights the benefit of leveraging the diffusion model representations for mask generation.

\input{tables/ablation}

\paragraph{Randomizing location or scale.}

To understand if object \emph{locations} or object \emph{scales} are more important, we perform two sets of experiments: one samples the object location according to our model, but samples the scale unconditionally from the empirical data distribution; the other samples a location uniformly at random, but samples the scale according to our model (conditioned on the location).
We find that the augmentation gain in each case is roughly half of our model's total, indicating that location and scale are equally relevant, and that both need to be realistic to achieve an optimal performance gain.

\paragraph{Combining our location model and X-Paste.}

The purpose of our location model is to allow placing objects in a realistic context within a given scene.
It is intuitive that good locations and scale matter here, as the inpainting model can take advantage of these and \eg create challenging occlusion scenarios, but it may perform less well if locations are unrealistic.
An open question is whether improving locations has a similar effect on cut-and-paste type approaches.
Since X-Paste~\cite{xpaste} pastes pre-generated objects into the frames ignoring the context entirely, the realism of their scale or location should not matter.
Surprisingly, we still see a small performance boost when combining our location model with X-Paste.
However, augmented frames still appear entirely unrealistic, and the performance gain may only be due to an improved location and scale bias of the detector.

\paragraph{Additional analyses}

In the supplementary material, we attempt to quantify the realism and diversity of augmented frames, where our method yields numbers that are comparable to adding completely new data.
We further analyze the effect of bounding box refinement, where we use the predicted instance masks to refine bounding boxes.
While negligible at lower IoU thresholds, this has a significant influence at high thresholds.
Finally, we show examples of failure modes, which occur when the depth or drivable space predictions are incorrect, or for more complex scene geometries.

%% file: tables/sota_nuimages.tex
\begin{table*}[tb]
\caption{
Augmenting Faster R-CNN object detection on nuImages. While all augmentation methods improve over the base model, our proposed data augmentation with layout augmentation outperforms the other methods by a significant margin. \enquote{Replacement} is similar to
~\citep{gen2det,kupyn2024dataset}.
% \ah{Should not we include additional baselines compared in GeoDiffusion SOTA (copied from their Table 3), i.e., LAMA, Taming, LostGAN, ReCo, GLIGEN?}
% \da{I don't think they are very interesting, they perform worse than the baseline, they seem like broken models to me.}
The bottom row shows the percentage of instances belonging to each category in real training frames.
}
\label{tab:sota_nuimages}
\centering
\def\arraystretch{1.2}
\resizebox{\textwidth}{!}{
\begin{tabular}{clccccccccccccc}
\toprule
&& Locations & mAP & car & truck & trailer & bus & const. & bicycle & motor. & ped. & cone & barrier\\
\midrule
& Baseline & - & 37.8 \textsuperscript{~~~~~~~} & 53.6 & 41.8 & 17.2 & 43.1 & 25.5 & 45.4 & 46.9 & 32.0 & 32.8 & 39.3\\
\rowcolor{Gray}\cellcolor{white}& Replacement & original & 38.5 \textsuperscript{\textcolor{darkgreen}{+0.7}} & \textbf{54.2} & 43.3 & 17.7 & 44.5 & 26.1 & 46.1 & 47.8 & \textbf{32.2} & \textbf{33.1} & \textbf{40.1}\\
& Random Loc. & random & 38.2 \textsuperscript{\textcolor{darkgreen}{+0.4}} & 53.4 & 42.8 & 15.8 & 44.6 & 27.0 & 46.4 & 48.4 & 31.5 & 32.6 & 39.4\\
\rowcolor{Gray}\cellcolor{white} & X-Paste~\cite{xpaste} & random & 38.2 \textsuperscript{\textcolor{darkgreen}{+0.4}} & 53.7 & 42.9 & 16.0 & 44.1 & 26.4 & 46.4 & 48.7 & 31.8 & 32.7 & 39.5\\ 
& GeoDiffusion~\cite{geodiffusion} & original & 38.3 \textsuperscript{\textcolor{darkgreen}{+0.5}} & 53.2 & 43.8 & 18.3 & 45.0 & 27.6 & 45.3 & 46.9 & 30.5 & 32.1 & 39.8\\

% \rowcolor{Gray}\cellcolor{white}& Background \cite{li2024simple} & original & 36.6 \textsuperscript{\textcolor{darkred}{-1.2}} & 53.0 & 41.9 & 13.6 & 42.1 & 23.9 & 43.7 & 46.3 & 30.0 & 31.7 & 39.4 \\

\rowcolor{Gray}\cellcolor{white}
\multirow{-6}{*}{\cellcolor{white}\rotatebox[origin=c]{90}{800$\times$456}}
& \textbf{Ours} & scene-aware & \textbf{39.2} \textsuperscript{\textcolor{darkgreen}{+1.4}} & 53.9 & \textbf{44.0} & \textbf{18.6} & \textbf{46.1} & \textbf{27.7} & \textbf{47.0} & \textbf{49.4} & 32.0 & 32.9 & 39.9\\
\midrule

& Baseline & - & 50.4 \textsuperscript{~~~~~~~} & 66.4 & 55.5 & 21.7 & 55.9 & 35.0 & 55.6 & 58.2 & 49.7 & 54.2 & 51.8\\
\rowcolor{Gray}\cellcolor{white} & Replacement & original  & 50.7 \textsuperscript{\textcolor{darkgreen}{+0.3}} & 66.8 & 55.4 & 22.9 & 56.5 & 34.7 & 55.3 & 58.9 & 49.6 & 54.3 & 52.1\\ 
& Random Loc. & random & 51.3  \textsuperscript{\textcolor{darkgreen}{+0.9}} & 66.4 & 56.3 & 23.0 & 58.0 & 36.4 & 56.8 & 60.2 & 49.7 & 54.8 & 51.3\\
\rowcolor{Gray}\cellcolor{white} & X-Paste~\cite{xpaste} & random & 51.5 \textsuperscript{\textcolor{darkgreen}{+1.1}} & \textbf{66.9} & 56.7 & 23.6 & 58.0 & 35.9 & 57.0 & 60.3 & \bfseries 50.0 & 54.7 & 52.2\\
& GeoDiffusion~\cite{geodiffusion} & original  & -- \textsuperscript{~~~~~~~} & -- & -- & -- & -- & -- & -- & -- & -- & -- & -- \\ 
\rowcolor{Gray}\cellcolor{white}
\multirow{-6}{*}{\cellcolor{white} \rotatebox[origin=c]{90}{1600$\times$900}}
& \textbf{Ours} & scene-aware & \textbf{52.0} \textsuperscript{\textcolor{darkgreen}{+1.6}} & \bfseries 66.9 & \bfseries 56.9 & \textbf{25.3} & \textbf{58.8} & \textbf{37.5} & \textbf{57.1} & \textbf{60.7} & 49.8 & \textbf{54.9} & \textbf{52.6}\\
\midrule

& Real data [\%] & & & 37.1 & 5.4 & 0.6 & 1.2 & 0.9 & 2.5 & 2.5 & 24.4 & 12.6 & 12.8 \\
\bottomrule

\end{tabular}
}
\end{table*}

%% file: tables/masks_nuimages.tex
\begin{table*}[tb]
\caption{
Augmenting Mask R-CNN instance segmentation and detection on nuImages at full $1600 \times 900$ resolution. All approaches use the same inpainting strategy and only differ by the locations in which objects are inpainted to the scene.}
\label{tab:masks_nuimages}
\centering
\def\arraystretch{1.2}
\resizebox{0.95\textwidth}{!}{
\begin{tabular}{lc|cccccc|ccccccc}
\toprule
& Locations &\multicolumn{6}{c|}{Bounding box evaluation} & \multicolumn{6}{c}{Instance mask evaluation}\\
&& mAP & mAP$_{50}$ & mAP$_{75}$ & small & med. & large
& mAP & mAP$_{50}$ & mAP$_{75}$ & small & med. & large\\
\midrule
% \multirow{3}{*}{1x}
% &Base 
% & xx.x & xx.x & xx.x & xx.x & xx.x & xx.x 
% & xx.x & xx.x & xx.x & xx.x & xx.x & xx.x
% \\
% &Repl. 
% & xx.x & xx.x & xx.x & xx.x & xx.x & xx.x 
% & xx.x & xx.x & xx.x & xx.x & xx.x & xx.x
% \\
% &Ours
% & xx.x & xx.x & xx.x & xx.x & xx.x & xx.x 
% & xx.x & xx.x & xx.x & xx.x & xx.x & xx.x
% \\
% \midrule
Baseline  & - 
& 51.2 \textsuperscript{~~~~~~~} & 77.8 & 55.8 & 31.3 & 49.5 & 64.2
& 41.5 \textsuperscript{~~~~~~~} & 71.2 & 42.2 & 19.6 & 40.5 & 57.6
\\
\rowcolor{Gray}
Replacement & original
& 51.5 \textsuperscript{\textcolor{darkgreen}{+0.3}} & 77.9 & 56.1 & \bfseries 31.4 & 50.0 & 64.6
& 41.6 \textsuperscript{\textcolor{darkgreen}{+0.1}} & 71.6 & 42.3 & \bfseries 19.7 & 40.8 & 57.8
\\
Random Loc. & random
& 51.9 \textsuperscript{\textcolor{darkgreen}{+0.7}} & 77.9 & 56.2 & 31.0 & 50.2 & 65.2
& 42.1 \textsuperscript{\textcolor{darkgreen}{+0.7}} & 71.9 & 42.9 & 19.4 & 40.9 & 58.3
\\
\rowcolor{Gray}
\bfseries Ours & scene-aware
& \bfseries 52.6 \textsuperscript{\textcolor{darkgreen}{+1.4}} & \bfseries 78.8 & \bfseries 57.0 & 31.1 & \bfseries 50.8 & \bfseries 66.1
& \bfseries 42.4 \textsuperscript{\textcolor{darkgreen}{+0.9}} & \bfseries 72.4 & \bfseries 43.1 & 19.4 & \bfseries 41.0 & \bfseries 58.9
\\
\bottomrule
\end{tabular}}
\end{table*}

%% file: tables/sota_bdd.tex
\begin{table*}[tb]
\caption{
Faster R-CNN object detection augmentation results on BDD100K at full $1280 \times 720$ resolution. 
}
\label{tab:sota_bdd}
\centering
\def\arraystretch{1.2}
\resizebox{0.95\textwidth}{!}{
\begin{tabular}{lccccccccccccc}
\toprule
& Locations & mAP & ped. & rider & car & truck & bus & train & motor. & bicycle & tr.light & tr.sign\\
\midrule
Baseline & -
& 31.4 \textsuperscript{~~~~~~~} & 34.5 & 26.3 & 50.8 & 46.2 & 46.9 & 0.0 & 24.6 & 25.9 & 21.9 & \bfseries 37.1 \\
\rowcolor{Gray} Replacement & original
& 31.6 \textsuperscript{\textcolor{darkgreen}{+0.2}} & 34.4 & 26.5 & 51.1 & 46.2 & 48.4 & 0.0 & 24.9 & 25.5 & 21.7 & 36.8 \\
Random Loc. & random
& 32.1 \textsuperscript{\textcolor{darkgreen}{+0.7}} & 34.6 & 27.2 & 51.0 & 47.3 & 48.9 & 0.0 & 25.9 & 26.8 & \bfseries 22.1 & 37.0\\
\rowcolor{Gray} X-Paste~\cite{xpaste} & random 
& 32.3 \textsuperscript{\textcolor{darkgreen}{+0.9}} & 34.8 & 27.5 & 50.9 & \bfseries 47.9 & 49.2 & \bfseries 3.4 & 24.9 & 26.5& 21.8 & 36.6 \\ 
\textbf{Ours} & scene-aware &
\textbf{32.7} \textsuperscript{\textcolor{darkgreen}{+1.3}} & \bfseries 35.0 & \bfseries 28.0 & \bfseries 51.2 & 47.6 & \bfseries 49.9 & 0.8 & \bfseries 27.6 & \bfseries 27.9 & 21.9 & \bfseries 37.1 \\
\bottomrule
\end{tabular}
}
\end{table*}

%% file: tables/ablation.tex
% \begin{table*}[b]
% \caption{Ablating design choices in our approach, using FasterRCNN for detection on nuImages. See text for more details.}
% \label{tab:ablation}
% \centering
% % \resizebox{\textwidth}{!}{
% \begin{tabular}{cccccccc}
% \toprule
% & \textcolor{mediumgray}{Baseline} & Ours & w/o ControlNet & Rand. loc. & w/o masks & SAM masks \\
% \midrule
% 800$\times$456 & \textcolor{mediumgray}{37.8} & 39.2 & 37.9 \textsuperscript{\textcolor{red}{-1.3}} & 38.3 \textsuperscript{\textcolor{red}{-0.9}} & 38.7 \textsuperscript{\textcolor{red}{-0.5}} & 39.0 \textsuperscript{\textcolor{red}{-0.2}} \\
% 1600$\times$900 & \textcolor{mediumgray}{50.4} & 51.9 & 50.8 \textsuperscript{\textcolor{red}{-1.1}} & 51.3 \textsuperscript{\textcolor{red}{-0.6}} & 51.4 \textsuperscript{\textcolor{red}{-0.5}} & 51.4 \textsuperscript{\textcolor{red}{-0.5}} \\
% \bottomrule
% \end{tabular}
% % }
% \end{table*}

\begin{table}
\caption{Ablation of the effect of design choices in our approach on Faster R-CNN detector performance (mAP) on nuImages.
%See text for more details.
}
\label{tab:ablation}
\small
\centering
\def\arraystretch{1.2}
% \resizebox{\linewidth}{!}{
\begin{tabular}{lcc}
\toprule
& 800$\times$456 & 1600$\times$900 \\
\midrule

\textcolor{mediumgray}{Baseline} & \textcolor{mediumgray}{37.8}\textsuperscript{~~~~~~~~} & \textcolor{mediumgray}{50.4}\textsuperscript{~~~~~~~~} \\

Ours & 39.2\textsuperscript{~~~~~~~~} & 52.0\textsuperscript{~~~~~~~~} \\

\rowcolor{Gray} \quad without finetuning & 37.9 \textsuperscript{\textcolor{red}{-1.3}} & 50.8 \textsuperscript{\textcolor{red}{-1.2}} \\

\quad without mask pred. & 38.7 \textsuperscript{\textcolor{red}{-0.5}} & 51.4 \textsuperscript{\textcolor{red}{-0.6}} \\

\rowcolor{Gray} \quad with SAM masks & 39.0 \textsuperscript{\textcolor{red}{-0.2}} & 51.4 \textsuperscript{\textcolor{red}{-0.6}} \\

\quad model loc., rand. scale & 38.6 \textsuperscript{\textcolor{red}{-0.6}} & 51.6 \textsuperscript{\textcolor{red}{-0.4}} \\

\rowcolor{Gray} \quad rand. loc., model scale & 38.5 \textsuperscript{\textcolor{red}{-0.7}} & 51.6 \textsuperscript{\textcolor{red}{-0.4}} \\

X-Paste~\cite{xpaste} & 38.2\textsuperscript{~~~~~~~~} & 51.5\textsuperscript{~~~~~~~~} \\

\rowcolor{Gray} \quad with our location model & 38.4 \textsuperscript{\textcolor{darkgreen}{+0.2}} & 51.6 \textsuperscript{\textcolor{darkgreen}{+0.1}} \\

\bottomrule
\end{tabular}
% }
\end{table}

%% file: sections/5_conclusion.tex
\section{Conclusions}

In this work, we demonstrate that generative data augmentation benefits from adding objects in new and realistic locations.
We first propose a scene-aware probabilistic location model that predicts new object locations for existing scenes.
To fully take advantage of this location model, we then adapt a diffusion model to jointly inpaint objects in the proposed locations and to produce instance masks for them.
Using this approach, we are able to generate realistic and challenging augmented frames, \eg with object occlusions, which set a new state of the art in data augmentation for object detectors on two street scene datasets, outperforming the mAP gains achieved by existing methods by a large margin.
We also demonstrate significant gains in data augmentation for instance segmentation.
Crucially, using the same augmentation strategy but with completely random object placement, or only reusing existing object locations, performs much worse than our approach, highlighting the benefit of augmentation that places objects in new and realistic locations.

\paragraph{Limitations.} The probabilistic factorization of our location model takes advantage of the high regularity of street scenes.
We suspect that more for diverse scenes, as for example in COCO~\cite{coco}, a fully learned location model may be required.
At the same time, we expect that a more advanced object placement strategy would improve performance even further.
Our location model is also somewhat tied to the augmentation strategy, it requires an inpainting model to fully take advantage of the predicted locations.
The benefit for cut-and-paste approaches is limited.
% , as approaches that add objects in new locations (XPaste, Ours) show the strongest improvement.
It also depends on the quality of the depth estimation and the segmentation of drivable space (we show failure cases in the Appendix).

An opportunity for future work is combining the augmentation approaches discussed in this work.
In particular, object replacement, full-frame synthesis, and our object placement in new locations are in principle complementary.
However, identifying the right way to combine synthetic data from these sources is a non-trivial problem~\cite{gen2det}.
Finally, we expect all generative data augmentation to improve with the quality of the underlying generator.
We test our approach using only Stable Diffusion 2~\cite{rombach2022stablediffusion}, as it is a commonly used open-source model, but other promising open-source or open-weight models have been released since~\cite{chen2024pixartsigma,flux2024}.
As generative models progress, generative data augmentation will likely play an increasingly important role in the training of task-specific models.

%%%%%%%%%%%%%%%%%%
% \paragraph{Impact statement}

% Generative data augmentation has the potential to substantially improve performance for under-represented classes in various perception tasks.
% In autonomous driving for example, it has the potential to increase detector robustness in rare situations, thereby preventing costly mistakes.
% Synthesizing hard to obtain scenarios and examples may also be a cost-effective strategy in real world use cases.

% At the same time, augmenting training data with generative models may overemphasize biases already present in the training data.
% A generative model might have inherent biases, and introducing the data it generates into the training pipeline of a downstream perception network can decrease its robustness, cause it to overfit, or have other adverse effects.
% It remains important to measure the quality of generated data before using it in the training pipeline, and to test its effect on the augmented perception networks, especially in safety-critical applications.

%% file: sections/X_supplement.tex
\clearpage
\setcounter{page}{1}
\appendix
\maketitlesupplementary

\section{Method Details}
\label{sec:appendix:method}

\subsection{Location model}
\label{sub:appendix:method:locationmodel}

In this section we provide more details in our location model, specifically the individual sampling steps and how we approximate the corresponding likelihoods.
The accompanying figure is \cref{fig:location_model_eval}, where we show \enquote{car}, \enquote{bus}, and \enquote{pedestrian} as representative classes, using data from the front camera in the nuImages dataset (nuImages uses 6 cameras, and we treat them separately).

\paragraph{Depth sampling}
We use DepthAnything~\cite{depthanything} as an off-the-shelf depth estimator.
While the model technically outputs what the authors call \emph{disparity} ($disparity\propto 1/depth$), we still refer to it as depth, as we believe our description is easier to understand this way.
\Cref{fig:location_model_eval:depth} shows depth histograms for the three classes, along with the log-normal approximation we use.
We prefer an easy-to-use parametric distribution over one that optimizes data fit, and in this case a log-normal is clearly a good enough choice. Note that larger depth values are closer to the camera, so most objects are comparatively far away.

\paragraph{Location sampling}
Once we have sampled a depth value, we select all pixels from the drivable space which are within $\tau_d=5$ of this value.
This typically results in a band of possible locations, two examples of which are shown in \cref{fig:location_model_eval:example}.
We then select a pixel from this band at random and use it as the bottom-center location for the bounding box.
Should no pixels in the drivable space be within the allowed depth interval, we reset the depth value to the closest allowed one.
By first sampling the depth, and only then a location from the resulting band, we avoid oversampling close objects, because logically there are more pixels closer to the camera than further away.

\paragraph{Height sampling}
Sampling the height is arguably the most complicated part in our model, as it is conditioned on the depth. \Cref{fig:location_model_eval:height_vs_depth} shows example histograms at different depths.
We also approximate these with log-normals, but the approximation is clearly not as good as in the case of the depth.
Nevertheless, we find that on average it results in realistic object heights.
To get the mean $\mu_h(d)$ and standard deviation $\sigma_h(d)$ of the log-normal for a given depth, we build such histograms for all possible disparities, evaluated in windows of width 2, and then calculate the parameters in each case (\ie the mean and standard deviation of the log-data).
We then fit a simple parametric model of the form $y=a+b\cdot x^c$ to be able to interpolate mean $\mu_h(d)$ and standard deviation $\sigma_h(d)$ for a given depth at sampling time.
The interpolation is visualized in \cref{fig:location_model_eval:height_param_fit} and fits the data fairly well.
Only at high depth values, \ie close to the camera, do we find significant deviation from the underlying data.
As these depths have very low likelihoods anyway, we accept this tradeoff.

\paragraph{Width sampling}
We sample the width conditioned on the height via the distribution of aspect ratios for the given class, visualized in \cref{fig:location_model_eval:aspect_ratios}.
These are independent of the depth.
Unfortunately, the aspect ratio histograms follow a more complex pattern, and we were unable to find a good parametric approximation.
This is likely due to objects living in 3D space with almost arbitrary rotations (in terms of yaw), whereas we only work with 2D bounding boxes.
This effect is very pronounced for cars and buses, but less so for pedestrians.
As a result, we use the empirical likelihoods directly, \ie the histogram bins are normalized to sum to 1 and then taken as likelihoods for the corresponding bin intervals.
\input{figures/M_multiscale_agg}
\input{figures/location_model_eval}
\subsection{Mask generation}
\label{sub:appendix:method:maskgeneration}
\input{figures/mask_decoder}
As mentioned in \cref{sub:method:gendataaug}, we design a simple mask decoding module $\mathcal{M}$ to plug into the SD2 inpainting model, responsible for creating segmentation masks for every generated object.
As explained in the main text, its architecture is mirroring the one of the image (VAE) decoder, with two notable exceptions.
First, all its layers feature 4x fewer channels, as we assume that decoding a binary mask requires a lot less capacity than to decode an RGB image.
Secondly, it does not use the clean latents $\textbf{z}_0$ directly, but rather multi-scale features from the encoder part of the SD2 UNet, generated by feeding $\textbf{z}_0$ to it.
This step is akin to running the SD2 denoiser for an additional diffusion step, and extracting representations from several layers in its UNet (specifically, all downsampling layers).
Given the fact that the input to the $\mathcal{M}$ is not a tensor but rather a set of tensors at different resolutions (which we name $\{ \mathbf{r}_1, \mathbf{r}_2, \dots, \mathbf{r}_d \}$), we run a simple multi-scale aggregation module (represented in \cref{fig:M_multiscale_agg}), which upscales all features to the same resolution and concatenates them.

We train our mask decoder on nuImages, as all its labeled objects come with an instance mask that we can use for supervision. 
We optimize $\mathcal{M}$ after the finetuning stage of SD2, and  observed no benefits in training both jointly.
As BDD100k does not provide any segmentation masks, we use the mask decoder trained on nuImages when generating objects for this dataset.
We observed that generated masks are of comparable quality, even for unseen classes that are not available in nuImages (such as `rider' and `train').

Besides providing pseudo-labels for instance segmentation augmentation, the masks decoded by the mask decoder prove useful in the case generated objects occlude themselves.
Consider, for instance, the case represented in \cref{fig:mask_decoder}, where instances of a construction vehicle and of a traffic cone are generated independently, in close locations.
When pasting the crops back to the original frame using full bounding box areas, it is impossible to avoid visible artifacts.
However, by using precise pixel masks, occlusions are handled successfully, increasing the overall realism of the augmented frame.
We believe such cases are one of the main factors explaining the drop in mAP that we observe when not using masks, as reported in \cref{tab:ablation} and \cref{fig:bbox_refinement}.

% More details on mask generation (architecture!), and how we avoid artifacts.

\section{Additional Results \& Examples}
\label{sec:appendix:results}

\subsection{Realism and diversity}
\label{sub:realism_diversity}
\input{figures/realism_diversity}
In this section, we evaluate and compare the realism and diversity of frames augmented with our method and other methods, as these aspects are often mentioned as key factors for good data augmentation performance \citep{dvornik2019importancevisualcontext, gen2det}.
Here, realism refers to the visual quality of augmented samples,
whereas diversity describes how different augmented frames are from the already available training frames.
We argue that the current methods relying on original or random object locations maximize one aspect while neglecting the other:
generating from original locations guarantees realistic scene composition, yet variations are limited to visual appearance, whereas using random locations yields very diverse scene layouts and scale for new objects, at the expense of realism.

This can be seen qualitatively in the examples in \cref{fig:cover}.
To measure the realism of augmented frames quantitatively, we adopt the approach of GeoDiffusion~\cite{geodiffusion} and evaluate a pretrained Mask R-CNN\footnote{The ImageNet-pretrained R-50 model from mmdetection3d~\cite{mmdet3d2020}.} on the augmented frames.
To measure diversity, we compute the FID between paired sets of unedited and edited \emph{full frames}, using \num{1000} images for both.
A method that does not change the data at all should obtain a FID score of zero, and a method that produces high diversity samples should obtain higher FID.
Importantly, generating unrealistic frames (\eg random noise) can result in very high FID scores: as such, we argue that the optimal value for this diversity metric is not necessarily the lowest or highest score, but rather a score that is comparable to that achieved by a different set of real frames.

\Cref{fig:realism_diversity} shows realism and diversity scores for several augmentation methods.
The replacement baseline achieves low frame FID and high mAP: the augmented frames are realistic, but hardly changed.
GeoDiffusion and X-Paste have much higher frame FID, meaning they add a lot of diversity, but this comes at the price of lower realism.
GeoDiffusion likely obtains high frame FID because it generates both background and objects.
Our approach changes parts of the scene but it leaves other areas of the frame untouched, and achieves high realism according to the pretrained detector.
Importantly, it achieves comparable frame FID to a set of \emph{Real} data, meaning our augmented frames are as dissimilar to the originals as a set of new real frames. 
We argue that this is a very good operating point in the realism-diversity tradeoff.
Finally, we see that the mAP of the random location baseline is similar to that of X-Paste (which also uses random locations), but with lower frame FID.
We suspect that this is because the inpainting model sometimes struggles to generate objects in unrealistic locations, and instead just completes the crop to look realistic.

\subsection{Effect of bounding box refinement}
\label{sub:appendix:results:bbox_refinement}

\input{figures/bbox_refinement}

We adapt the inpainting model so that it also predicts instance masks for the generated objects.
We use these masks to refine the bounding boxes, so that they fit the generated objects more closely, which leads to a significant performance improvement (see ablations in \cref{tab:ablation}).
As one might expect, this bounding box refinement has a stronger influence when the the IoU threshold for a successful detection is higher.
In \cref{fig:bbox_refinement} we show the mAP of an augmented Faster R-CNN on nuImages (full resolution) for different IoU thresholds.
Up to a threshold of around 0.75 there is hardly a difference between refined and normal bounding box use.
But at higher thresholds the effect becomes more apparent, with an improvement of 23$\%$ at the highest threshold 0.95.

\subsection{Effect of finetuning}
\label{sub:appendix:results:finetuning}

\input{figures/ours_vs_sdinpaint}

\input{tables/ablation_appendix}

We choose to finetune the SD2 inpainting model~\cite{rombach2022stablediffusion}, for which we tried both direct finetuning and ControlNet~\cite{controlnet} and observed similar performance (see \cref{tab:appendix:ablations}, results in this work are from ControlNet).
While we expect that other methods of finetuning~\cite{lora,t2i} work similarly well, we showed in the ablations that finetuning in general has a strong effect on augmentation performance.
Examples of the main differences we observe are shown in \cref{fig:ours_vs_sdinpaint}.
In many cases, with finetuning the generated objects visually fit their context better, \ie they have more realistic colors and saturation (top row in \cref{fig:ours_vs_sdinpaint}).
We also find that in a significant number of cases, the non-finetuned model doesn't produce an object at all, instead just completing the area to look realistic (middle row in \cref{fig:ours_vs_sdinpaint}).
In other cases, the non-finetuned model does produce an object, but it only fills part of the provided bounding box, whereas the finetuned model tends to fill the desired box almost fully (bottom row in \cref{fig:ours_vs_sdinpaint}).

% \subsection{X-Paste with our location model}
% \label{sec:appendix:xpaste_ours}

% \input{tables/xpaste_ours}

% In \cref{tab:appendix:xpaste_ours} we show the performance of X-Paste\cite{xpaste} when combined with our location model.
% There is only a small gain with respect to the random locations used by X-Paste, compared to a large difference between our approach and the \enquote{Replacement} approach shown in the main paper.
% We suspect that this is in large part due to the fact that X-Paste does not consider the scene layout to begin with.
% Objects are always pasted on top of the underlying frame, so they will still not blend in realistically, regardless of the quality of our proposed locations.
% It is also impossible to model realistic occlusions when using X-Paste.

\subsection{Further ablations \& baselines}
\label{sub:appendix:further_ablations}

\input{figures/background_augmentation}

In \cref{tab:appendix:ablations} we show the performance of background augmentation~\cite{li2024simple} and compare the ControlNet-finetuned (default) and directly finetuned versions of our approach.
Using our approach with ControlNet and direct finetuning lead to the same overall mAP, with only small differences in individual classes.
This suggest that the specific method of finetuning is not important, only that finetuning is performed at all.
As there is no official code release for background augmentation, we reimplemented the method ourselves.
While we are confident in our implementation, there is a possiblity that we missed some important detail, leading to poor performance.
With augmented data from this method, mAP is reduce by $1.2$ points.
Another possible explanation is a data mismatch between the model and the nuImages dataset---the method produces convincing results on some frames and fails completely on others (see \cref{fig:appendix:background_augmentation}).
We suspect that with finetuning or quality filtering the approach could be improved significantly.

\subsection{Failure cases}
\label{sub:appendix:results:failurecases}

\input{figures/additional_examples_failure}

We could identify some failure cases in our augmentation approach, examples of which are shown in \cref{fig:failure_cases}.

The first is a result of inpainting with instance masks.
In most cases, our mask decoder predicts object masks that don't include shadows.
Depending on the lighting in the original scene, this can give our objects a \enquote{floating} appearance, which hurts visual realism (top row in \cref{fig:failure_cases}).
The reason for this behaviour is that the instance masks in the original data, which our mask decoder is trained with, do not include shadows either.
As a result, we were unable to find a way to test if this actually hurts augmentation performance or only visual realism.

The second failure case occurs when the depth estimate for the scene is wrong.
We use DepthAnything~\cite{depthanything}, which produces relative depth estimates (or disparity to be more precise, see \cref{sec:appendix:method}) that are normalized to the distance value of the most distant parts of the scene.
This is usually the sky, but if there is no sky in the scene, \eg when the vehicle faces a wall, the depth is normalized to a closer value, which in turn leads to our model producing objects that are smaller than they should be (middle row in \cref{fig:failure_cases}).
We experimented with a version of DepthAnything finetuned for metric depth estimation, but still observed the same behaviour.

Finally, our location model relies on segmentation of drivable space to place objects. Mistakes in the segmentation map will sometimes lead object placement that is visually unrealistic. In the example in \cref{fig:failure_cases} (bottom row), the segmentation model identifies both the grass and the barrier above it as \enquote{terrain}, so that the silver car is rendered in a location where it doesn't fit geometrically.
This issue should disappear with better segmentation models, but we were unable to test how this influences augmentation performance.

\subsection{Qualitative examples}
\label{sub:appendix:results:qualexamples}

We show more examples of augmented frames for all methods compared in this work.
Examples for ours are shown in \cref{fig:additional_examples_ours}, for object replacement in \cref{fig:additional_examples_replace}, for random placement in \cref{fig:additional_examples_random}, for X-Paste~\cite{xpaste} in \cref{fig:additional_examples_xpaste}, and for GeoDiffusion~\cite{geodiffusion} in \cref{fig:additional_examples_geodiffusion}.

\input{figures/additional_examples_ours}
\input{figures/additional_examples_replace}
\input{figures/additional_examples_random}
\input{figures/additional_examples_xpaste}
\input{figures/additional_examples_geodiffusion}

%% file: figures/M_multiscale_agg.tex
\begin{figure}[b]
    \centering
    \includegraphics[width=\columnwidth]{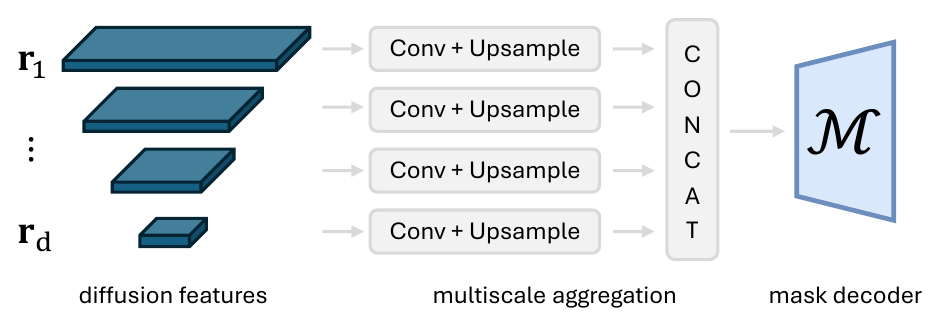}
    \caption{Representation of the multiscale aggregation module for feeding multi-scale UNet features to the proposed mask decoder.}
    \label{fig:M_multiscale_agg}
\end{figure}

%% file: figures/location_model_eval.tex
\begin{figure*}

    \centering
    \begin{subfigure}[b]{0.8\textwidth}
        \centering
        \includegraphics[width=\textwidth]{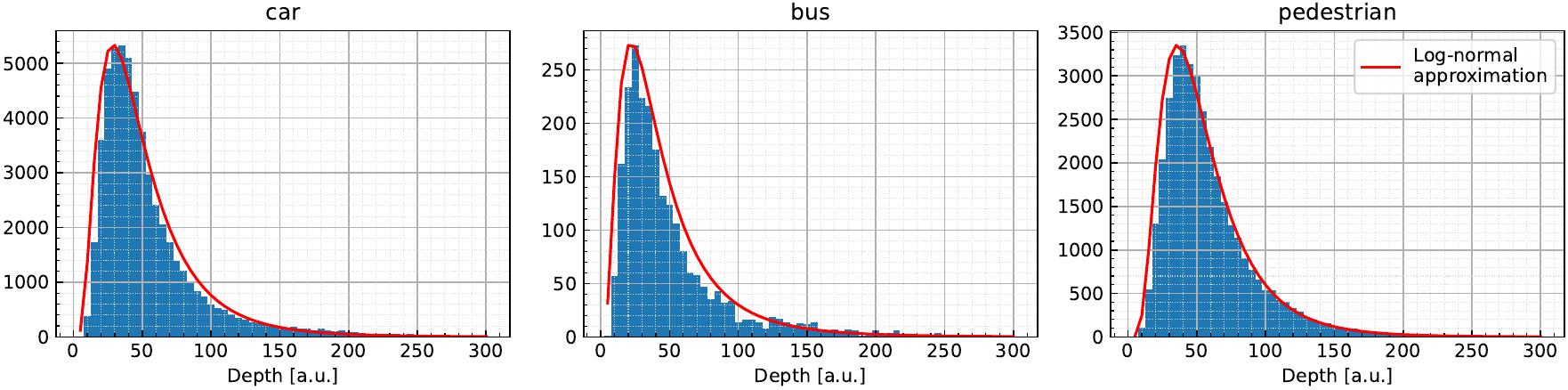}
        \caption{Depth histogram for three representative classes, along with the log-normal approximation we use. Note that the depth estimator we use outputs values where higher means closer to the camera.}
        \label{fig:location_model_eval:depth}
    \end{subfigure}

    \vspace{0.2cm}
    
    \begin{subfigure}[b]{0.8\textwidth}
        \centering
        \includegraphics[width=\textwidth]{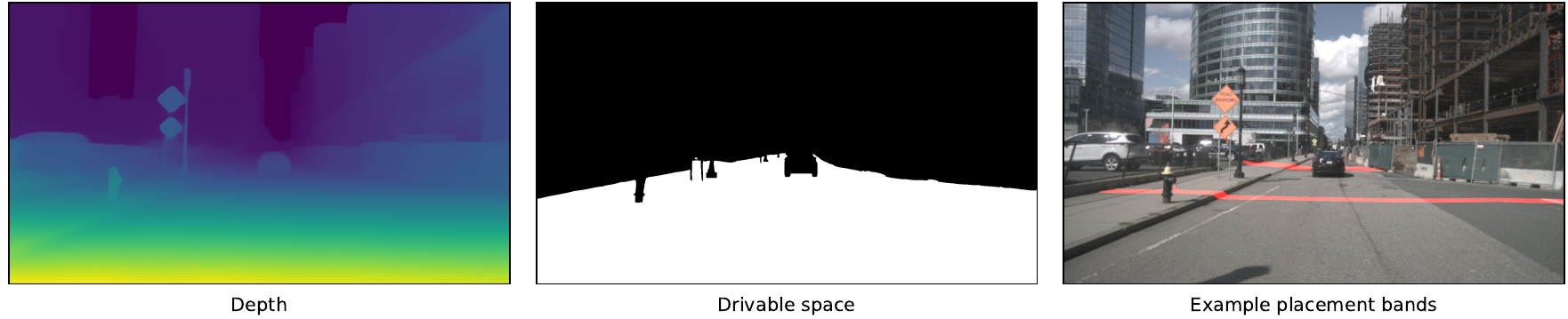}
        \caption{Example frame with depth map and drivable space. We first draw a depth value and then select an area from the drivable space that is within a threshold $\tau_d=5$ around that value. This results in placement bands from which a location is selected at random.}
        \label{fig:location_model_eval:example}
    \end{subfigure}

    \vspace{0.2cm}

    \begin{subfigure}[b]{0.8\textwidth}
        \centering
        \includegraphics[width=\textwidth]{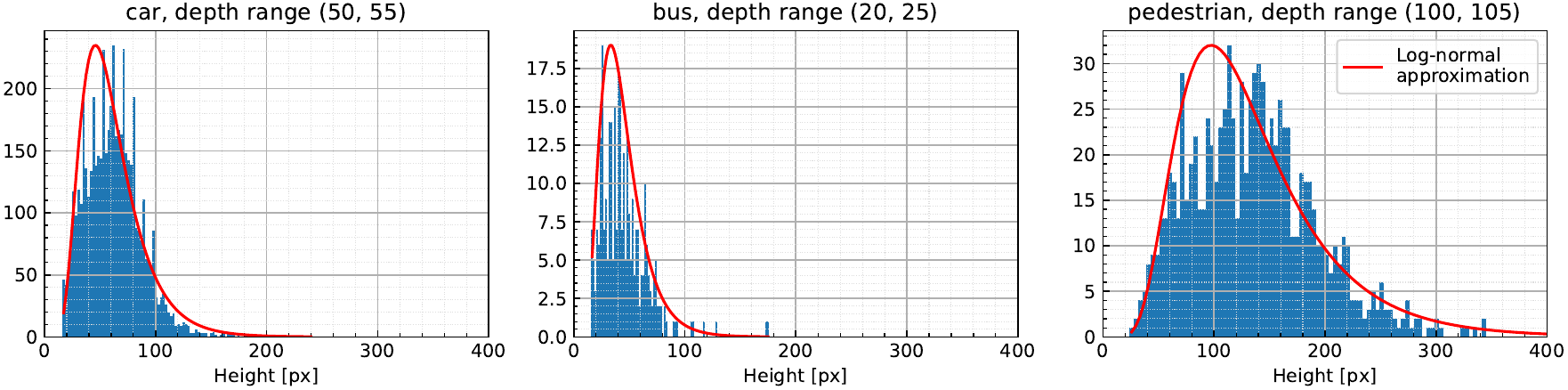}
        \caption{Height histogram for three representative classes, and three example depth ranges. We also approximate the resulting height distributions with log-normals.}
        \label{fig:location_model_eval:height_vs_depth}
    \end{subfigure}

    \vspace{0.2cm}

    \begin{subfigure}[b]{0.8\textwidth}
        \centering
        \includegraphics[width=\textwidth]{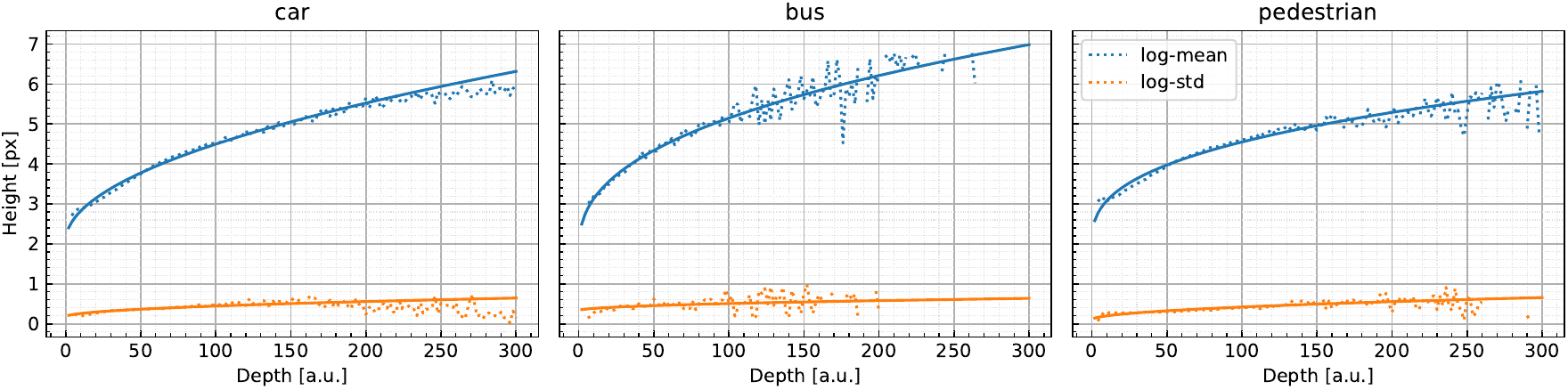}
        \caption{To be able to parametrize a height distribution, given a depth value, we compute mean and standard deviation of height histograms for different depths in the dataset (using a window size of 2). We then fit a curve to approximate $\mu_h(d)$ and $\sigma_h(d)$.}
        \label{fig:location_model_eval:height_param_fit}
    \end{subfigure}

    \vspace{0.2cm}

    \begin{subfigure}[b]{0.8\textwidth}
        \centering
        \includegraphics[width=\textwidth]{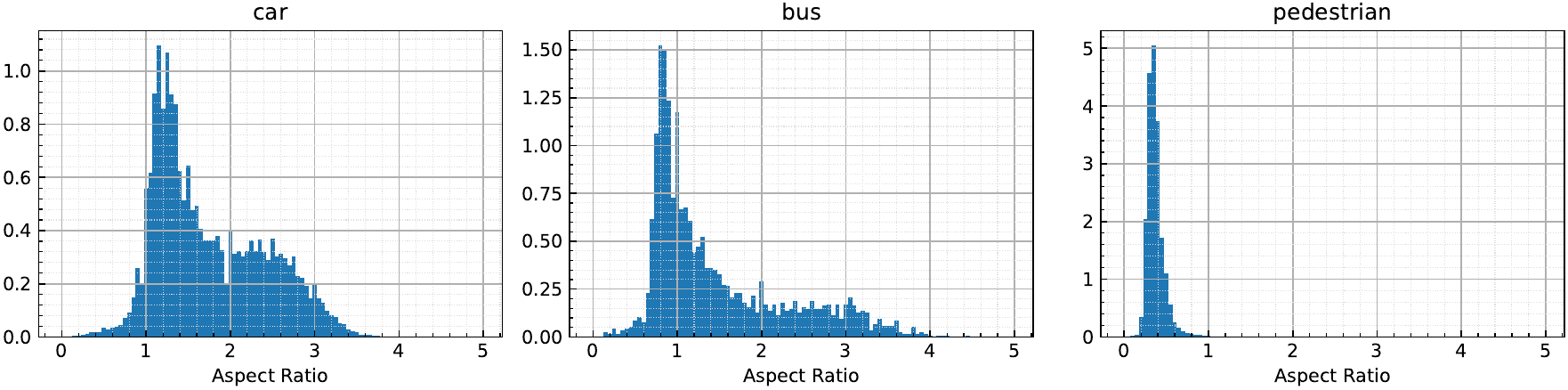}
        \caption{We get the object width for a given height via the aspect ratio. Because we could not find a satisfactory parametric distribution, we work with the empirical distributions directly, in this case.}
        \label{fig:location_model_eval:aspect_ratios}
    \end{subfigure}
     
    \caption{Overview of the different sampling steps in our location model, and what approximations we use.}
    \label{fig:location_model_eval}
    
\end{figure*}

%% file: figures/mask_decoder.tex
\begin{figure*}[bth]
\centering
\includegraphics[width=1\linewidth]{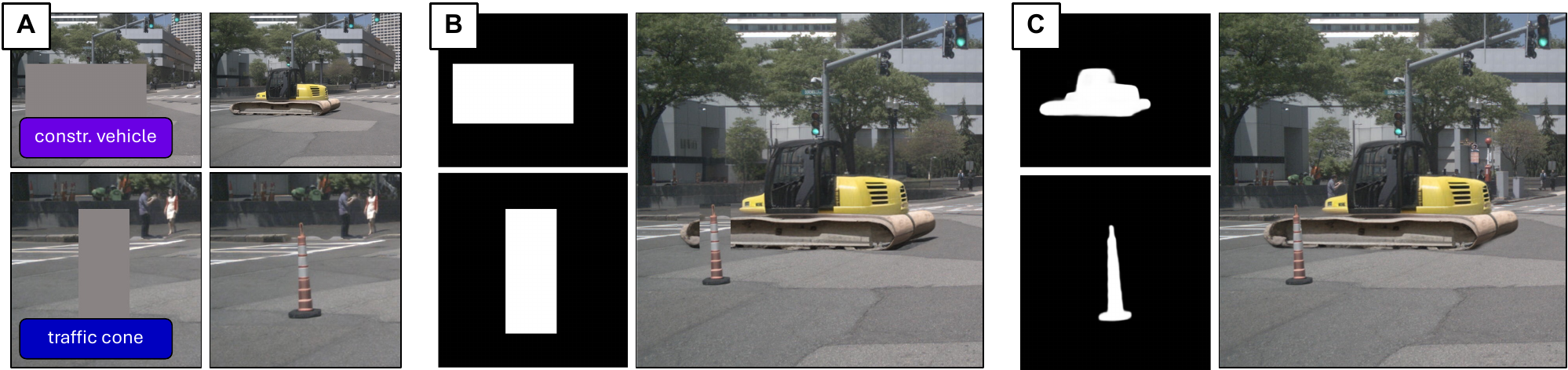}
\caption{
Producing object masks is crucial for seamless inpainting.
(A) Two example generations and their original inpainted areas.
(B) Pasting the synthetic objects back in the frame by using bounding boxes as masks can result in severe boundary artifacts, especially in the case of occlusions.
(C) Using generated instance masks yields more realistic results.
}
\label{fig:mask_decoder}
\end{figure*}

%% file: figures/realism_diversity.tex
\begin{figure}[t]
\centering
\includegraphics[width=0.95\columnwidth]{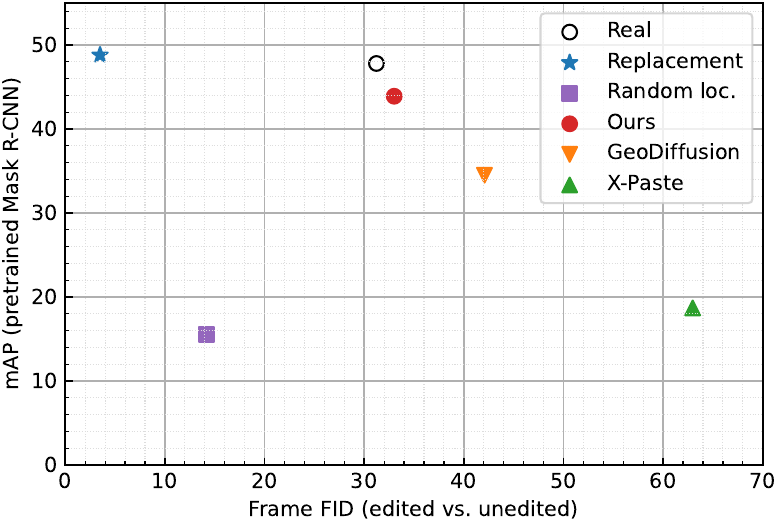}
\caption{Pretrained MaskRCNN mAP vs. FID between edited and unedited frames. High mAP indicates high realism, high frame FID indicates high frame variability.}
\label{fig:realism_diversity}
\end{figure}

%% file: figures/bbox_refinement.tex
\begin{figure}[t]
    \centering
    \includegraphics[width=\linewidth]{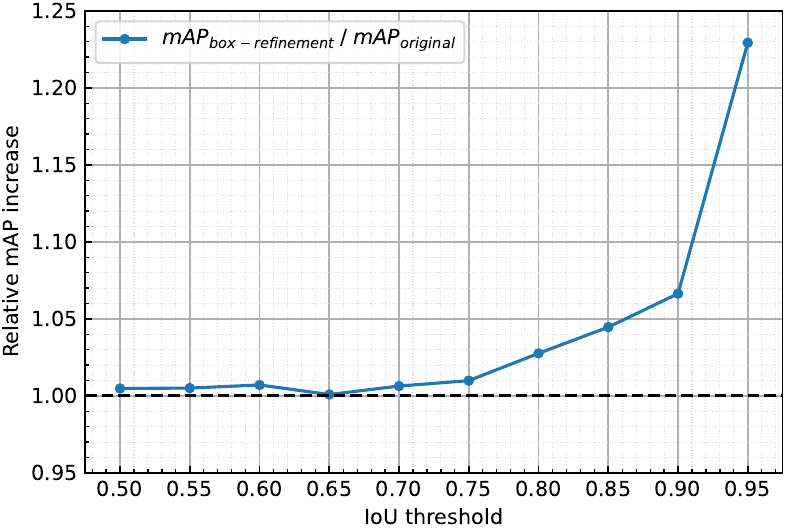}
    \caption{We use predicted instance masks to refine bounding boxes of inpainted objects. The graph shows the relative mAP improvement (on nuImages, full resolution) of this step, compared to not performing it, for different IoU thresholds. Up to a threshold of ca. 0.75 there is hardly a difference, but at higher thresholds this refinement results in up to 23$\%$ improvement.}
    \label{fig:bbox_refinement}
\end{figure}

%% file: figures/ours_vs_sdinpaint.tex
\begin{figure}[t]
    \centering
    \includegraphics[width=\linewidth]{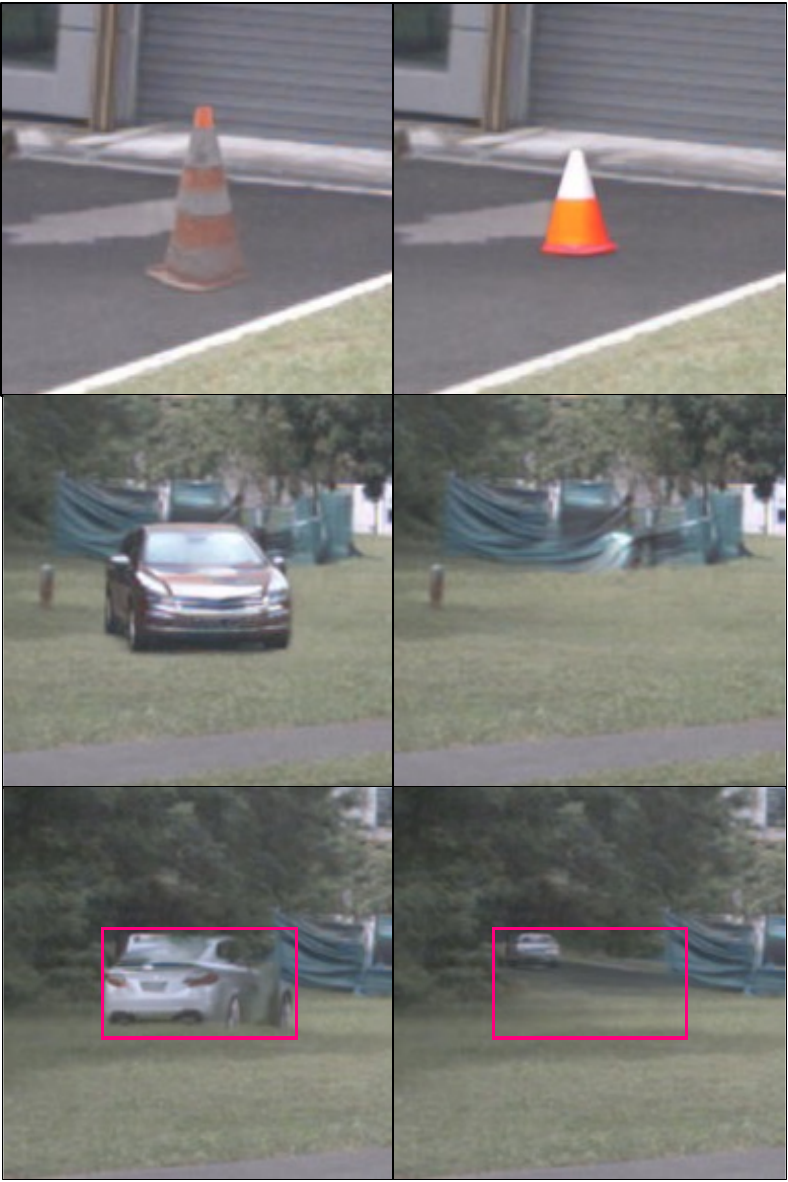}
    \caption{Comparison of finetuning with ControlNet (ours, left column) and using the SD2 pretrained inpainting checkpoint (right column). With finetuning, generated objects will often blend better with the surroundings in terms of color and saturation (top row). Without finetuning, the generator will sometimes not produce and object at all (middle row), or one that doesn't fully fill the provided bounding box (bottom row).}
    \label{fig:ours_vs_sdinpaint}
\end{figure}

%% file: tables/ablation_appendix.tex
\begin{table*}[t]
\caption{Further results for Faster R-CNN augmentation on nuImages ($800\times456$). As there is no public code available for \cite{li2024simple}, we reimplemented the method.}
\label{tab:appendix:ablations}
\centering
\def\arraystretch{1.2}
\resizebox{0.95\textwidth}{!}{
\begin{tabular}{lccccccccccccc}
\toprule
& Locations & mAP & car & truck & trailer & bus & const. & bicycle & motor. & ped. & cone & barrier\\
\midrule
Baseline & - & 37.8 \textsuperscript{~~~~~~~} & 53.6 & 41.8 & 17.2 & 43.1 & 25.5 & 45.4 & 46.9 & 32.0 & 32.8 & 39.3\\
\rowcolor{Gray} Background \cite{li2024simple} & original & 36.6 \textsuperscript{\textcolor{darkred}{-1.2}} & 53.0 & 41.9 & 13.6 & 42.1 & 23.9 & 43.7 & 46.3 & 30.0 & 31.7 & 39.4 \\
Ours (ControlNet) & scene-aware & 39.2 \textsuperscript{\textcolor{darkgreen}{+1.4}} & 53.9 & 44.0 & 18.6 & 46.1 & 27.7 & 47.0 & 49.4 & 32.0 & 32.9 & 39.9\\
\rowcolor{Gray} Ours (direct FT) & scene-aware & 39.2 \textsuperscript{\textcolor{darkgreen}{+1.4}} & 54.1 & 44.2 & 19.5 & 45.7 & 27.4 & 46.9 & 49.5 & 31.7 & 32.8 & 39.8\\
\bottomrule

\end{tabular}
}
\end{table*}

%% file: figures/background_augmentation.tex
\begin{figure}[t]
    \centering
    \includegraphics[width=\linewidth]{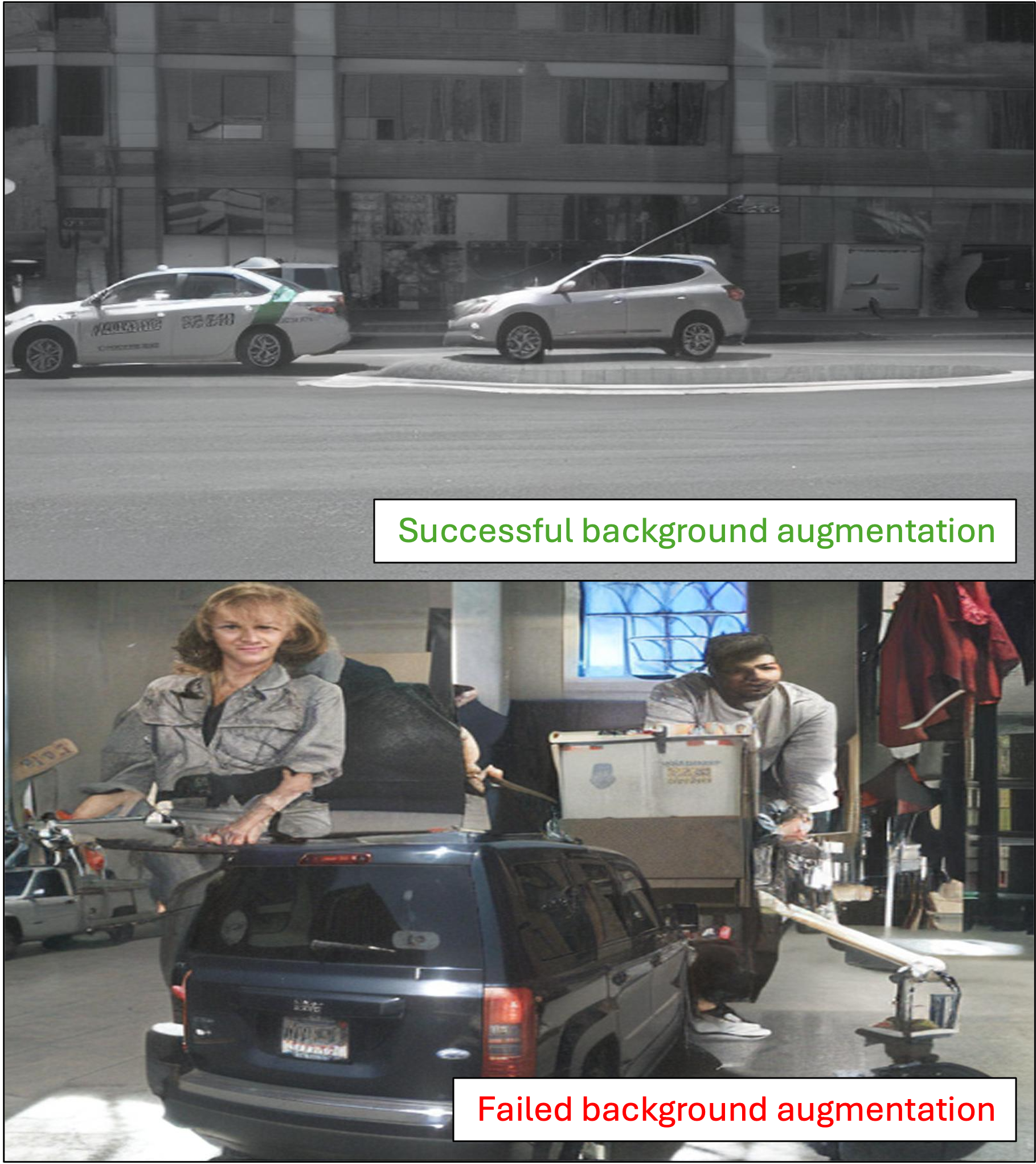}
    \caption{Examples of background augmentation~\cite{li2024simple}.}
    \label{fig:appendix:background_augmentation}
\end{figure}

%% file: figures/additional_examples_failure.tex
\begin{figure}[t]
    \centering
    \includegraphics[width=\linewidth]{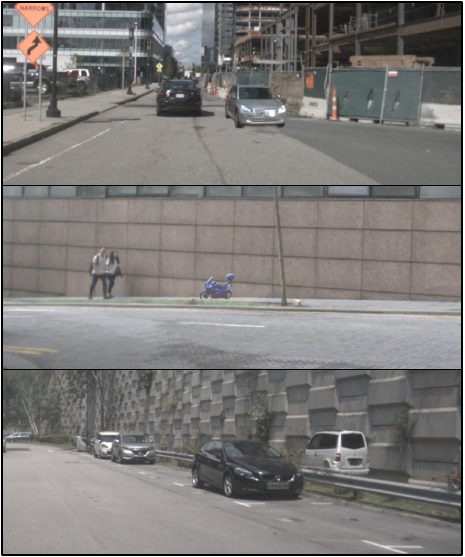}
    \caption{Some failure cases that occur in our augmentation approach. (top) Our mask decoder often fails to include shadows in the mask. (middle) If the depth estimate is wrong, our model produces objects with the wrong scale. (bottom) If the drivable space segmentation is wrong, our model puts objects in unrealistic locations.}
    \label{fig:failure_cases}
\end{figure}

%% file: figures/additional_examples_ours.tex
\begin{figure*}[t]
    \centering
    \includegraphics[width=\linewidth]{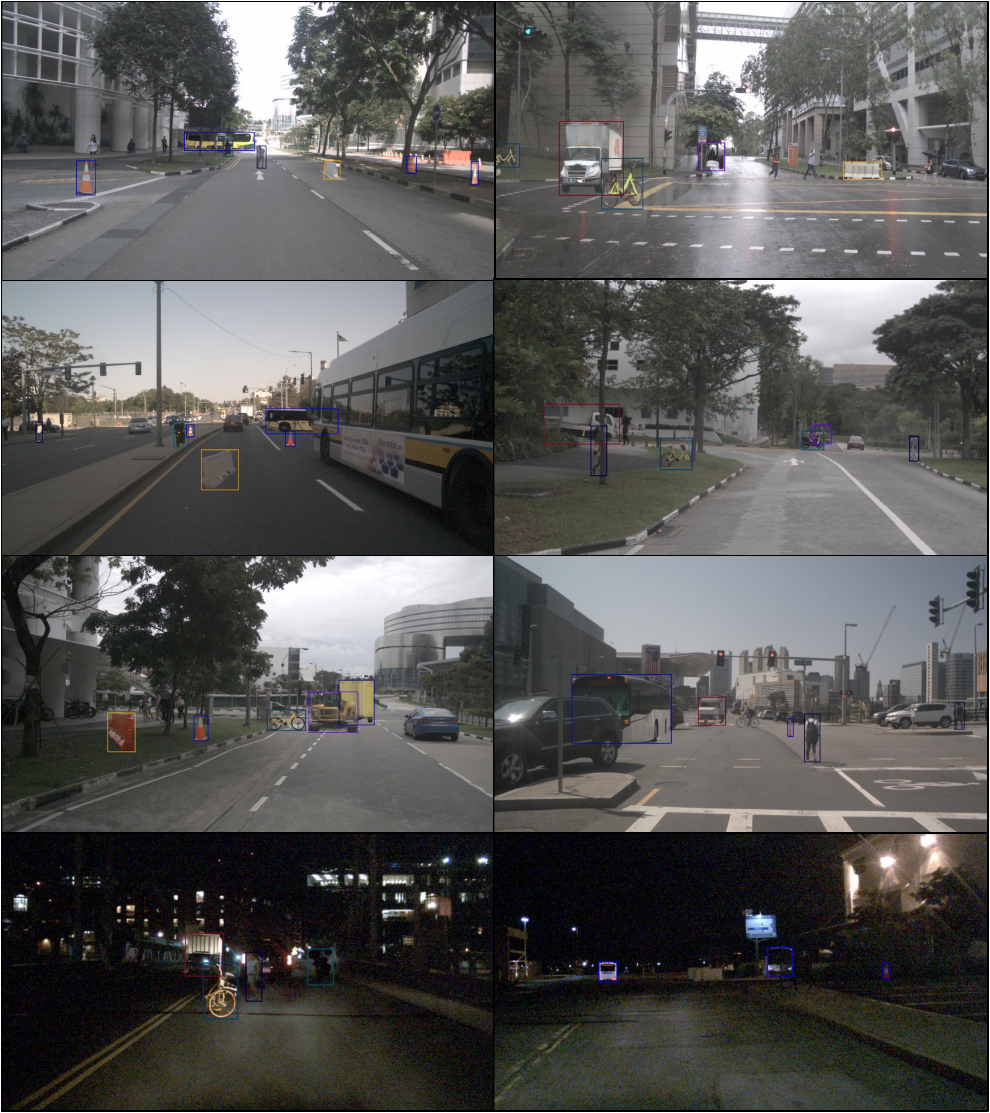}
    \caption{Additional frames augmented with our approach. We only show annotations for generated objects. Some frames are zoomed in for better visibility.}
    \label{fig:additional_examples_ours}
\end{figure*}

%% file: figures/additional_examples_replace.tex
\begin{figure*}[t]
    \centering
    \includegraphics[width=\linewidth]{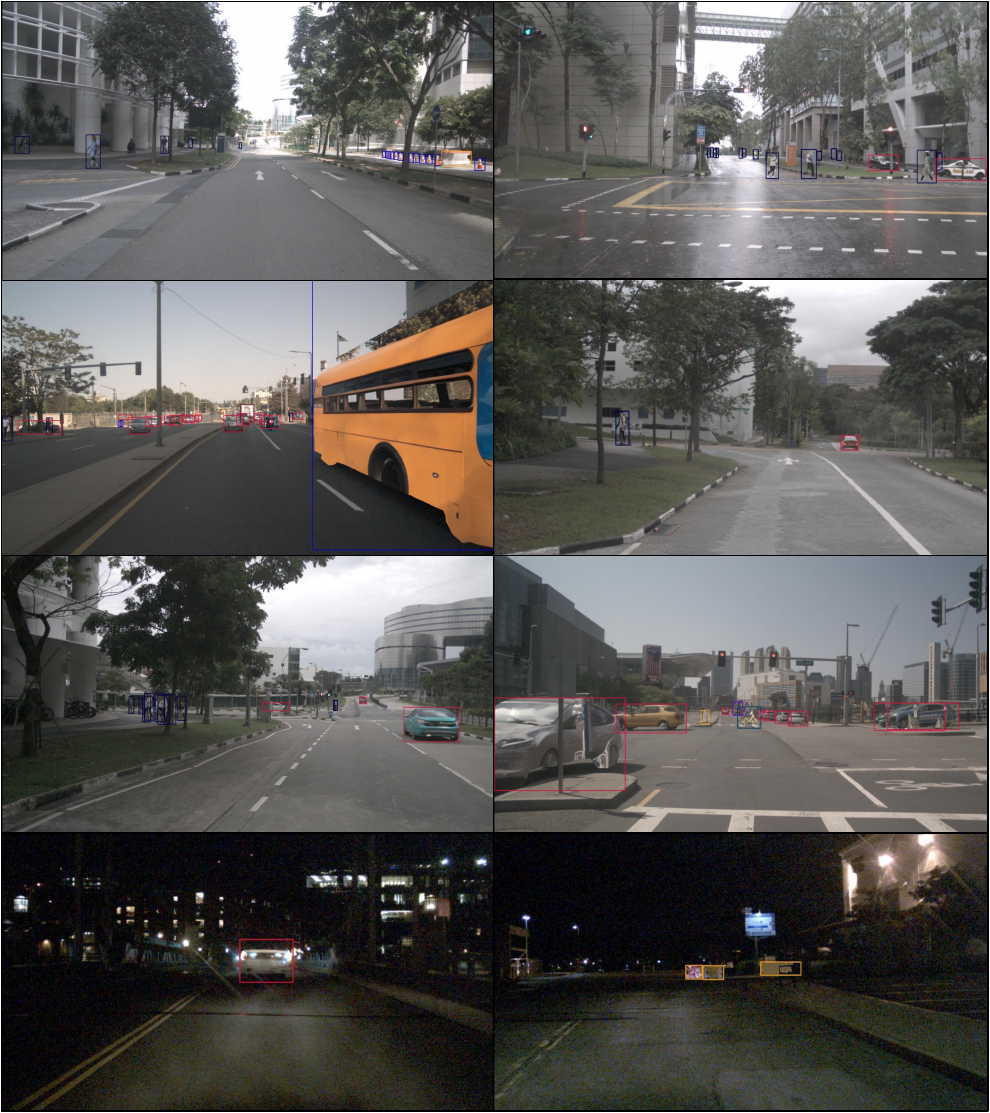}
    \caption{Additional frames augmented with object replacement. We only show annotations for generated objects. Zoom factors are chosen to match \cref{fig:additional_examples_ours}.}
    \label{fig:additional_examples_replace}
\end{figure*}

%% file: figures/additional_examples_random.tex
\begin{figure*}[t]
    \centering
    \includegraphics[width=\linewidth]{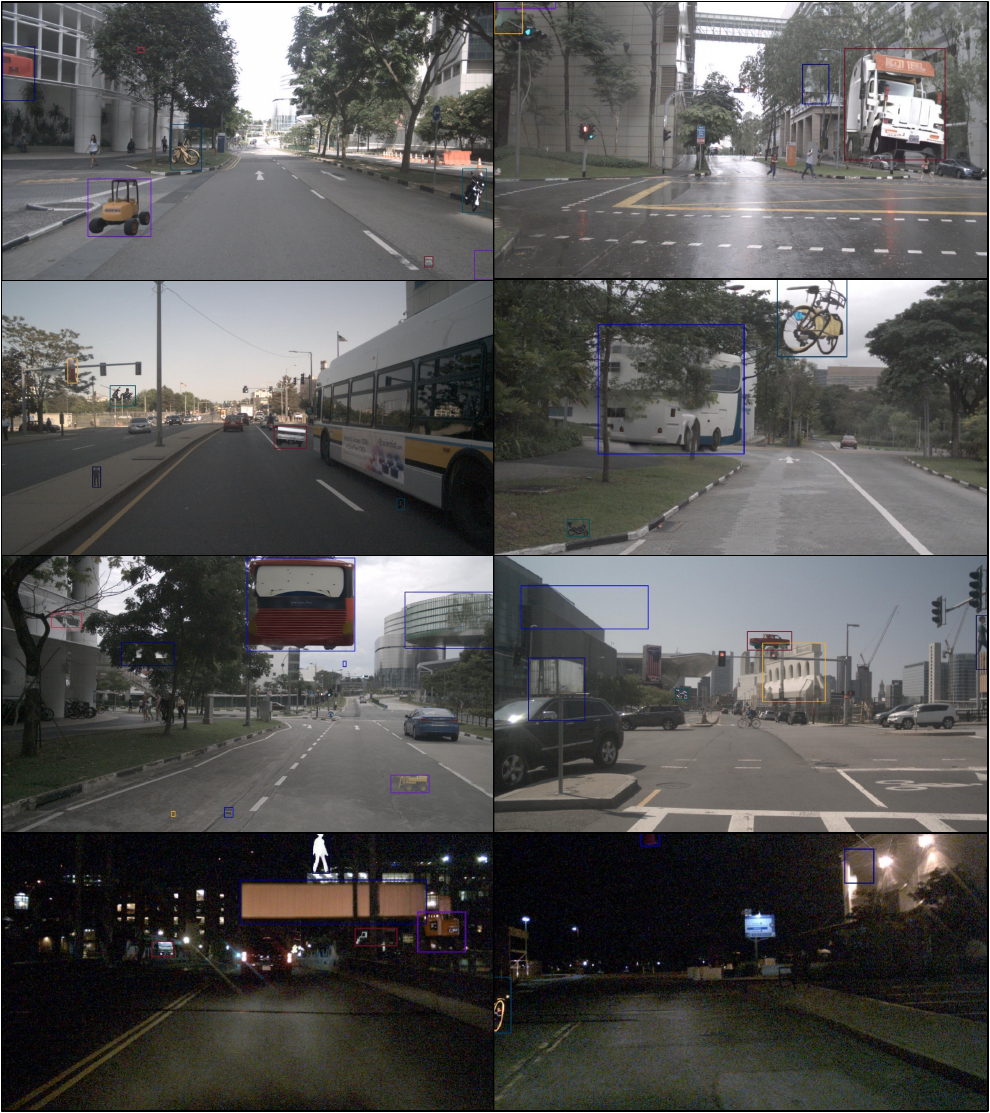}
    \caption{Additional frames augmented with random placement. We only show annotations for generated objects. Zoom factors are chosen to match \cref{fig:additional_examples_ours}.}
    \label{fig:additional_examples_random}
\end{figure*}

%% file: figures/additional_examples_xpaste.tex
\begin{figure*}[t]
    \centering
    \includegraphics[width=\linewidth]{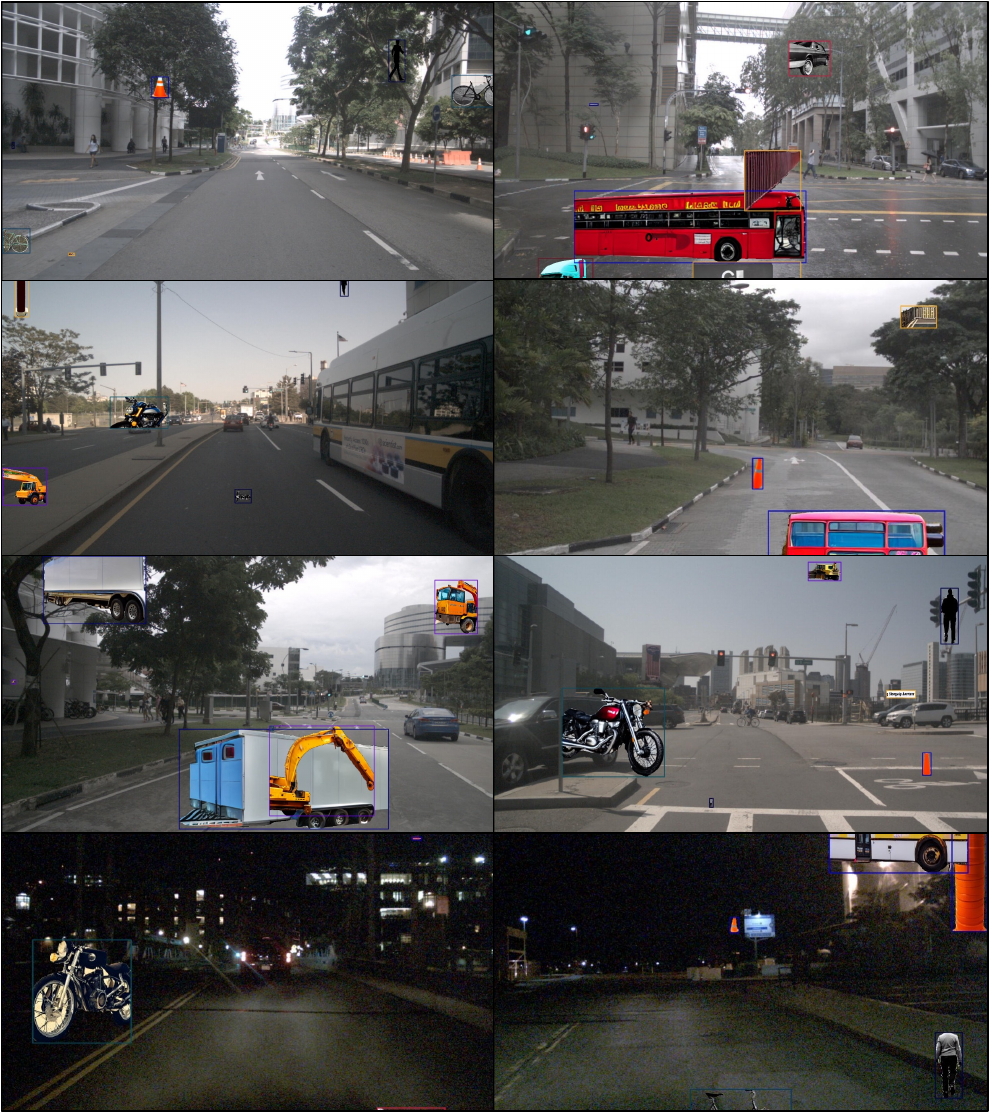}
    \caption{Additional frames augmented with X-Paste~\cite{xpaste}. We only show annotations for generated objects. Zoom factors are chosen to match \cref{fig:additional_examples_ours}.}
    \label{fig:additional_examples_xpaste}
\end{figure*}

%% file: figures/additional_examples_geodiffusion.tex
\begin{figure*}[t]
    \centering
    \includegraphics[width=\linewidth]{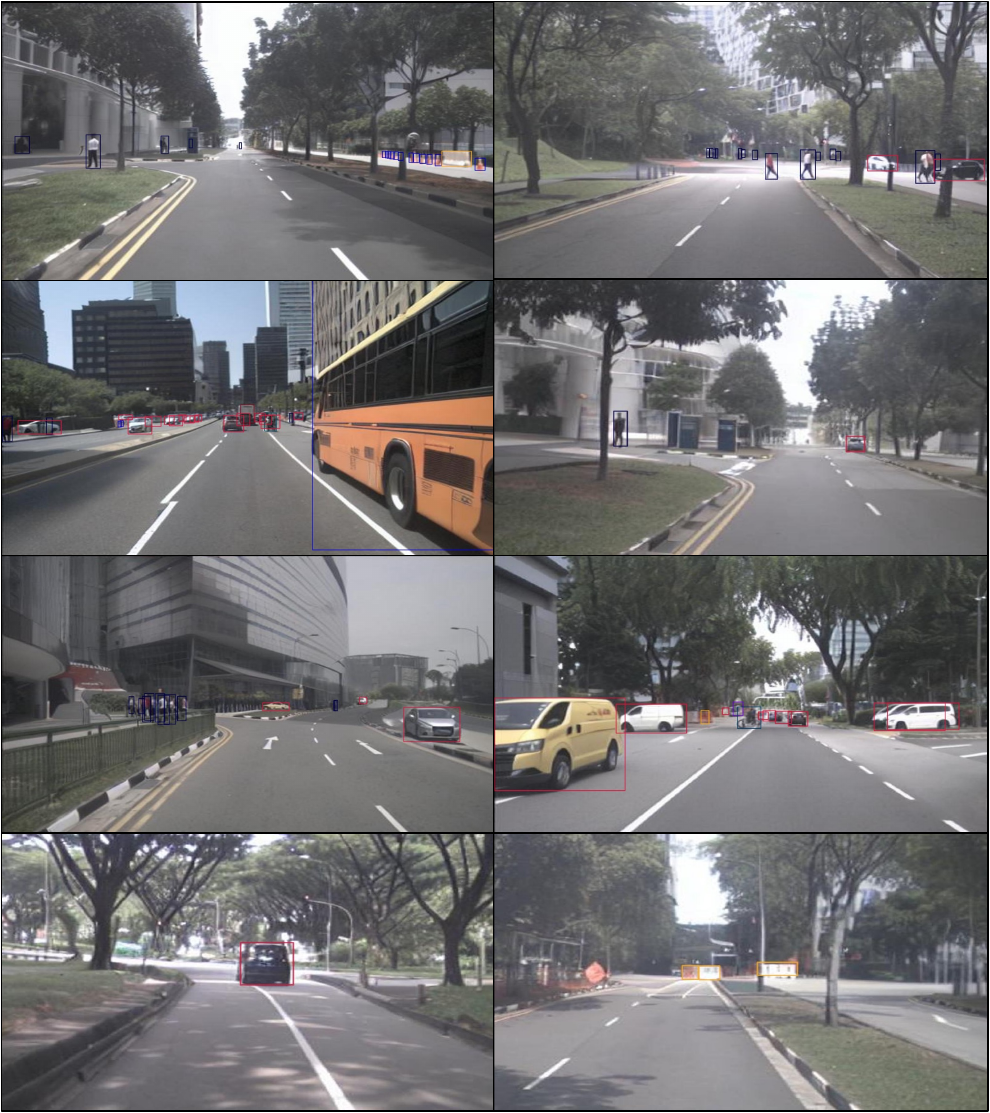}
    \caption{Additional frames augmented with GeoDiffusion~\cite{geodiffusion}. Zoom factors are chosen to match \cref{fig:additional_examples_ours}.}
    \label{fig:additional_examples_geodiffusion}
\end{figure*}